\documentclass[runningheads]{llncs}

\usepackage{eccv}

\usepackage{eccvabbrv}

\usepackage{graphicx}
\usepackage{booktabs}
\usepackage{wrapfig}

\usepackage[accsupp]{axessibility}  

\newcommand*\samethanks[1][\value{footnote}]{\footnotemark[#1]}

\usepackage{hyperref}

\usepackage{orcidlink}

\begin{document}

\title{Revisit Human-Scene Interaction via Space Occupancy} 


\author{
Xinpeng Liu\inst{1}\orcidlink{0000-0002-7525-3243}\thanks{Equal contribution.} \and
Haowen Hou\inst{1,2}\orcidlink{0009-0002-6944-0939}\samethanks \and
Yanchao Yang\inst{3}\orcidlink{0000-0002-2447-7917} \and
Yong-Lu Li\inst{1}\orcidlink{0000-0003-0478-0692}\thanks{Corresponding author.}  \and
Cewu Lu\inst{1}\orcidlink{0000-0003-1533-8576}
} 

\authorrunning{Liu et al.}

\institute{Shanghai Jiao Tong University \and
Soochow University \and
The University of Hong Kong \\
\email{xinpengliu0907@gmail.com, haowenhou@outlook.com, yanchaoy@hku.hk, \{yonglu\_li, lucewu\}@sjtu.edu.cn}}

\maketitle

\begin{abstract}
Human-scene Interaction (HSI) generation is a challenging task and crucial for various downstream tasks.
However, one of the major obstacles is its limited data scale. 
High-quality data with simultaneously captured human and 3D environments is hard to acquire, resulting in limited data diversity and complexity. 
In this work, we argue that interaction with a scene is essentially interacting with the space occupancy of the scene from an \textit{abstract physical perspective}, leading us to a unified novel view of Human-Occupancy Interaction.
By treating pure motion sequences as records of humans interacting with invisible scene occupancy, we can aggregate motion-only data into a large-scale paired human-occupancy interaction database: Motion Occupancy Base (MOB).
Thus, the need for costly paired motion-scene datasets with high-quality scene scans can be substantially alleviated.
With this new unified view of Human-Occupancy interaction, a single motion controller is proposed to reach the target state given the surrounding occupancy.
Once trained on MOB with complex occupancy layout, which is stringent to human movements, the controller could handle cramped scenes and generalize well to general scenes with limited complexity like regular living rooms.
With no GT 3D scenes for training, our method can generate realistic and stable HSI motions in diverse scenarios, including both \textit{static} and \textit{dynamic} scenes.
The project is available at \url{https://foruck.github.io/occu-page/}.

\end{abstract}
    
\section{Introduction}
\label{sec:intro}
\begin{figure}[!t]
    \centering
    \includegraphics[width=\linewidth]{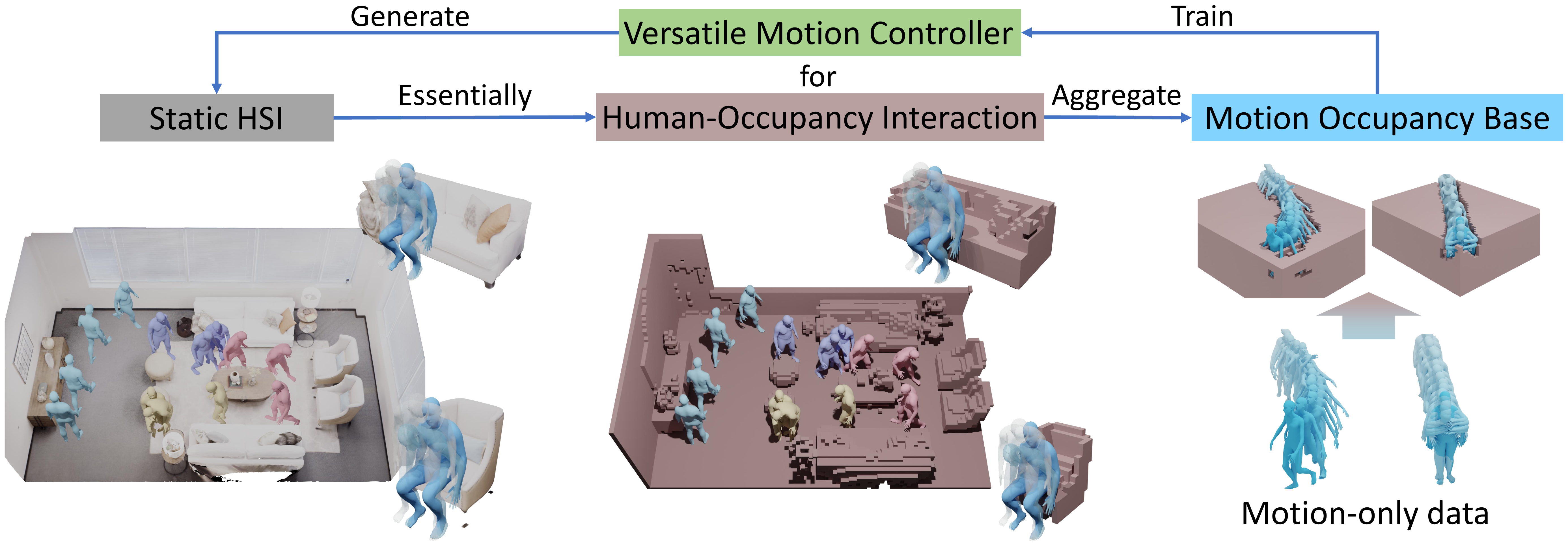}
    \captionof{figure}{
    We propose that interacting with scenes is essentially interacting with its space occupancy for static HSI. 
    In this view, we can unify motion-only data into a unified human-occupancy knowledge base and train a versatile Human-Occupancy Interaction controller upon it, achieving stable generation under various scenarios. 
    } 
    \label{fig:1}
\end{figure}
Human-Scene Interaction (HSI) generation has been an active field recently with its application in animation~\cite{starke2019neural} and embodied AI~\cite{puig2023habitat3,szot2021habitat}. 
For ideal HSI generation, models should be able to control the flexible articulated human body to conduct various motions in highly complex 3D environments, considering geometry constraints, motion feasibility, and diversity.
In this paper, we focus on interactions \textbf{without changing the scene}, which we refer to as \textbf{static HSI}.

The \textbf{limited data acquisition} is a major obstacle for static HSI generation.
Simultaneous capture of human and 3D scenes is expensive.
Some~\cite{samp,couch} collected data only for a narrow range of target objects.
Widely adopted PROX~\cite{prox} contains only 60 clips with limited diversity and complexity.
HUMANISE~\cite{humanise} synthesized data by fitting motion from large-scale motion base AMASS~\cite{amass} into various scenes.
CIRCLE~\cite{circle} put the scenes in VR applications to reduce the cost.
Most efforts take explicit scenes as necessary assets. 
Instead, we dig into the essence of static HSI and revisit it from a space occupancy view.

Space occupancy is usually identified as an elementary constraint for HSI: humans and scenes cannot occupy the same space simultaneously, which prevents humans from penetrating scenes.
We argue that occupancy is inherently the primary factor, especially for static HSI, which is essentially interaction with the scene occupancy.
As in Fig.~\ref{fig:1}, sitting on a chair is practically sitting on a chair-shaped \textit{solid being}.
The chair's occupancy allows us to rest our weight upon it, offering options to sit or stand, \ie, \textbf{affordances}, while its height and shape naturally invite sitting.
Other elements like category labels (\eg, identifying it as a ``chair'' semantically) influence our likelihood of choosing to sit.
The scene occupancy, therefore, offers a range of affordances and delineates potential interaction types, establishing an \textbf{initial} probability distribution for interactions. 
This is primarily accessible to the model and should be learned well at first. 

In light of this, the requirements of realistic scenes for HSI learning could be loosened, inspiring us to revisit large-scale motion datasets.
Instead of treating motion-only data as locomotion knowledge bases like existing efforts~\cite{zhang2022wanderings,zhao2023synthesizing}, we transform them into a comprehensive Human-Occupancy Interaction database, named Motion Occupancy Base (MOB).
For any given motion sequence, we first identify the space occupied by the human at any time in the sequence. 
The remaining space is conceptualized as the ``scene''. 
In this framework, the motion becomes a record of interaction with these defined space occupancies.
As a result, MOB aggregates enormous paired human-occupancy data from existing large-scale motion-only data.
Compared to current datasets~\cite{prox,circle,humanise}, MOB presents \textbf{over-complex} and \textbf{stringent geometric constraints}, pushing the model to the limit in interacting with space occupancy.

How do we exploit the aggregated MOB for static HSI?
Previous approaches generally provided two different answers.
Some~\cite{gimo,humanise,circle,huang2023diffusion} adopt an end-to-end pipeline from scene to motion, which is straightforward but inflexible without the ability to interact with dynamic scenes.
Some~\cite{wang2021synthesizing,wang2022towards,zhao2023synthesizing} typically decouple HSI into interaction and locomotion, adopting a two-stage paradigm: first generating key interaction frames and then synthesizing locomotion between them, with the help of methods like path-finding algorithms.
However, this decomposition overlooks the integrated nature of locomotion and interaction, like simultaneous arm movement and walking to navigate obstacles.
Our approach, rooted in space occupancy, reconceptualizes locomotion and interaction as a unified Human-Occupancy Interaction. We develop an auto-regressive Human-Occupancy Interaction controller for motion generation.
The controller processes various inputs, including surrounding occupancy and target locations, as control signals at each time step, producing future motion predictions.
Additionally, we introduce a field regulation module for enhanced collision avoidance.
There are three major advantages of our controller.
\textbf{First}, by learning from the challenging cases in MOB, it could handle complex scenes but also generalize well to regular simpler scenarios. 
\textbf{Second}, its versatility accommodates any static HSI without the need to switch between models for locomotion and interactions, even for \textit{HOI motion} synthesis. 
\textbf{Third}, its auto-regressive nature allows it to adapt to \textit{dynamic} environments such as an automatic door, a capability lacked in previous methods.
In extensive experiments, our controller performs effectively across various settings, even without training on real 3D scene data.

Overall, our contributions are three-fold:
1) We exhume Human-Occupancy Interaction as a major component of static HSI.
2) By digging out the inherent Human-Occupancy relationships of pure motions, we convert motion-only data into a paired Human-Occupancy database MOB, expanding the data scale.
3) We develop a versatile controller for flexible and stable HSI motion generation. 
\section{Related Works}
\label{sec:formatting}

{\bf Human-Scene Interaction Generation.}
Generating natural human motion sequences has received increasing attention with the emerging MoCap datasets~\cite{amass,kit,uestc,hml3d,humman,lin2023motion}.
Early efforts were paid to prefix/suffix-conditioned motion generation~\cite{hernandez2019human,harvey2020robust,athanasiou2022teach,guo2023back}, known as motion prediction and interpolation.
There also have been efforts to pursue generating motion with respect to specific action semantics, either represented in action categories~\cite{actor,guo2020action2motion,ActFormer} or natural language~\cite{temos,hml3d,tevet2022motionclip,motiondiffuse,hmdm,ReMoDiffuse,Petrovich_2023_ICCV,posegpt,t2mgpt,mld}.
Audio-conditioned motion generation was also explored, like music for dance generation~\cite{aist,li2021ai}.
However, these approaches tend to focus on human motion only, without placing the generated motion in a grounded 3D environment. 
Recently, progress has been made to advance HSI generation. 
Early efforts~\cite{li2020detailed,zhang2020perceiving} first reconstructed human-object interaction in 3D.
Some~\cite{zhang2020place,zhang2020generating,hassan2021populating} proceeded by placing SMPL-X~\cite{smplx} meshes in scanned 3D scenes. 
Semantic controls were also introduced~\cite{zhao2022compositional}.
For HSI motion generation, some~\cite{starke2019neural,samp,couch} focused on interacting with a single target object, mostly chairs.
GOAL~\cite{taheri2022goal} and SAGA~\cite{wu2022saga} generated full-body pre-grasp motions.
There have also been efforts~\cite{InterPhysHassan2023,tessler2023calm} for the physical control of specific interactions.
With paired motion-scene datasets~\cite{prox,circle}, approaches that synthesize HSI motion in cluttered scenes appeared.
Two-stage based algorithms~\cite{wang2021synthesizing,wang2022towards} were proposed to first generate static key-frame poses, and then interpolate them for a coherent motion.
GAMMA~\cite{zhang2022wanderings} proposed an RL framework for locomotion on flat grounds with the help of off-the-shelf path-finding algorithms.
DIMOS~\cite{zhao2023synthesizing} extended GAMMA with extra interaction policies.
In contrast, the follow-ups~\cite{gimo,humanise,circle,huang2023diffusion} adopted a simple end-to-end one-stage paradigm, while focusing more on boosting data.
GIMO~\cite{gimo} collected data with gaze information.
HUMANISE~\cite{humanise} synthesized pseudo-data from large-scale motion bases and 3D scan bases.
CIRCLE~\cite{circle} collected data with a VR application.
LAMA~\cite{lee2023locomotion} coupled the RL framework with a motion matching algorithm, generated HSI motions with only a unified policy, and leveraged motion editing for optimization.
Our major difference from previous efforts is the Human-Occupancy Interaction view, which alleviates the data hunger, and also enables a unified controller.
Moreover, we are the first to generate motion in a dynamic scene, which previous efforts failed to handle.

{\bf Scene Creation from Motion.}
Pose2Room~\cite{nie2022pose2room} predicted 3D object bounding boxes from human motion.
MOVER~\cite{yi2022human} reconstructed 3D scenes from videos that record human-scene interaction.
SUMMON~\cite{ye2022scene} and MIME~\cite{Yi_2023_CVPR} placed 3D object models concerning human motion.
These methods could be adopted to boost HSI data, though usually limited to a small range of indoor furniture.
\section{Motion Occupancy Base}
HSI generation could be formulated as $M=\mathcal{G}(S)$, where $M$ is the generated motion, $\mathcal{G}$ is the motion generator, and $S$ is a 3D scene. 
Conventionally, the scene is expected to be as realistic as possible, requiring expensive acquisition.
We find that these scenes could function almost equally to their occupancy for static HSI.
Thus, high-quality scenes might become less necessary here.
Given this, we revisit the large-scale motion-only datasets.
Intuitively, motion only contains body movement information.
But if we switch our sight to the vacant space around the moving person, considering the space as \textit{another form of occupancy}, we could find that these motions possess information on how humans interact with certain space occupancies.
To this end, we aggregate motion-only datasets into the human-occupancy interaction database: Motion Occupancy Base (MOB).

\begin{wrapfigure}{r}{0.5\textwidth}
    \centering
    \includegraphics[width=\linewidth]{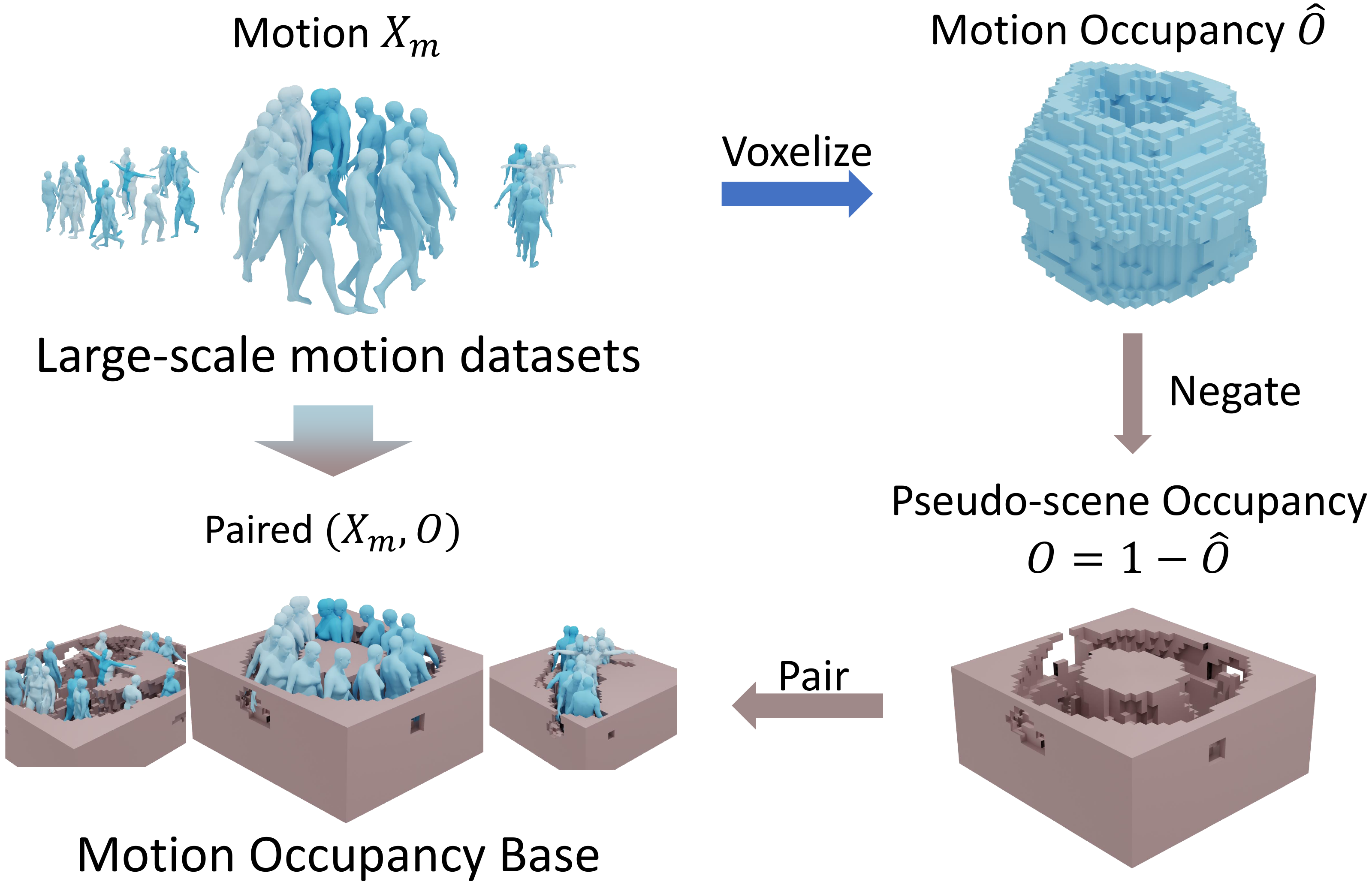}
    \caption{The construction process of Motion Occupancy Base. Note that the ceilings of the occupancy are hidden for clarity.}
    \label{fig:mob} 
\end{wrapfigure}

The overall pipeline is shown in Fig.~\ref{fig:mob}. 
Motion with SMPL(-X)~\cite{smpl,smplx} representation are collected, re-sampled to 30 FPS, and rotated so that the gravity direction points to the negative z-axis.
For motion sequence $X=\{x^i\}_{i=1}^t$, where $x_i$ is the SMPL(-X) representation in the $i$-th frame, $t$ is the number of frames, we first obtain mesh sequence $X_m=\{x_m^i\}_{i=1}^t$ via SMPL(-X) model, where $x_m^i=(f^i, v^i)$ is the mesh for the $i$-th frame, $f^i,v^i$ are the mesh faces and surface
vertices. A bounding box of $X_m$ is obtained based on vertices $V=\{v_i\}_{i=1}^t$.
Given the predefined grid unit size $u$, the box region is voxelized into motion occupancy volume $\hat{O}\in\{0,1\}^{\{H\times W\times D\}}$, computed in two steps: 1) $V$ occupy the corresponding voxels in $\hat{O}$.
2) Given the centers of the un-occupied voxels as $C=\{c_i\}_{i=1}^{|C|}$, if $c_i$ is inside any human mesh $x_m^i$, the corresponding voxel is occupied.
The pseudo-scene occupancy $O$ is then obtained by negating the motion occupancy, formulated as $O=1-\hat{O}$.
Thus, we convert motion-only sample $X$ into a motion-occupancy pair $(X, O)$ in MOB.
MOB aggregates 13 datasets with 98k instances.
More details are in the supplementary.

As shown in Fig.~\ref{fig:mob}, humans in MOB are acting with rather strict geometric constraints, which is uncommon in regular scenes covered by existing datasets~\cite{prox,circle,humanise}.
However, we show that by learning from hard cases in MOB, our model could grow its ability to handle both complex and regular scenarios well.

\section{Versatile Motion Controller}

\begin{figure}[!t]
    \centering
    \includegraphics[width=.9\linewidth]{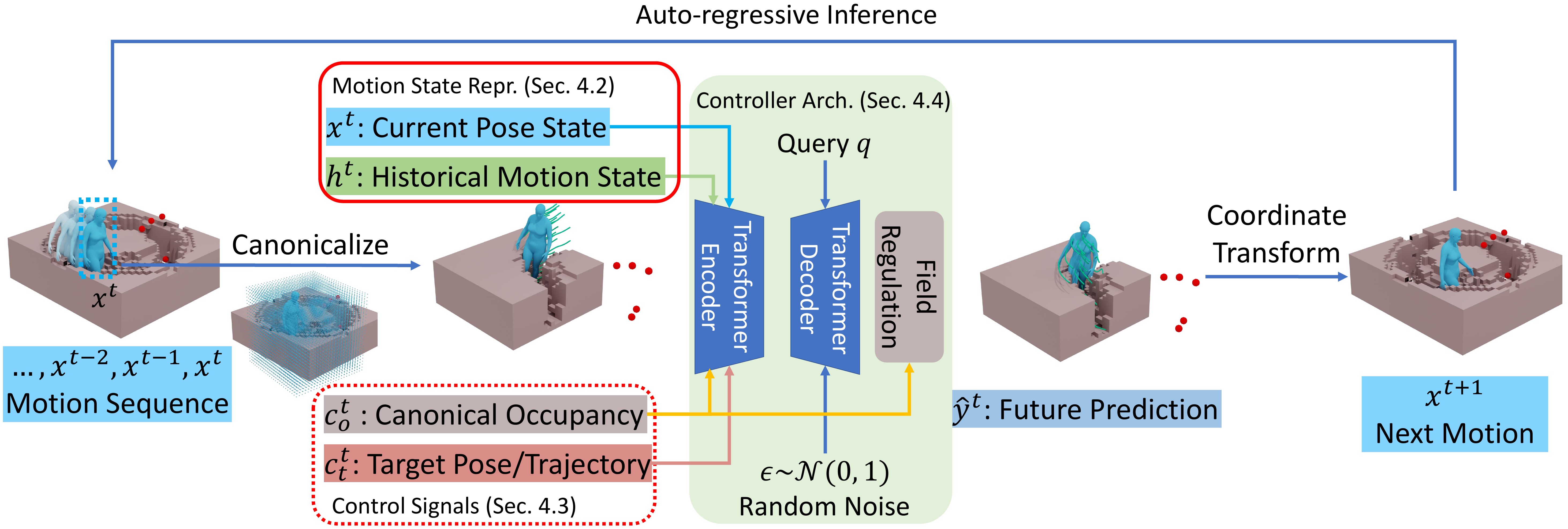}
    \caption{The architecture of our versatile motion controller. Given motion state and histories $x^t,h^t$, and control signals $c_o^t, c_t^t$, the controller auto-regressively generates the next motion frame w.r.t. the target pose and the canonical occupancy.}
    \label{fig:controller}
\end{figure}

\subsection{Preliminaries}
To instantiate $M=\mathcal{G}(S)$ for HSI generation, previous efforts either adopt a one-shot transformer generator or decompose $\mathcal{G}$ into two separate generators, one for interaction key-frame generation, the other for key-frame interpolation.
Instead, we formulate our generator as an auto-regressive controller.
At time step $t$, given historical motion state $h^t$, current pose state $x^t$, we expect a controller as 
\begin{equation}
    \hat{y}^t = \mathbf{G}(h^t, x^t, c^t),
\end{equation}
where $\hat{y}^t$ is the predicted future motion, and $c^t$ is the control signals. 
An overview is in Fig.~\ref{fig:controller}.
Sec.~\ref{sec:motion-state} introduces motion state representation $h^t,x^t,y^t$.
The control signals are covered in Sec.~\ref{sec:signals}.
The controller architecture is detailed in Sec.~\ref{sec:architecture}.
Sec.~\ref{sec:train-infer} demonstrates the training and inference procedure.

\subsection{Motion State Representation}
\label{sec:motion-state}

{\bf Notations.} 
In the global coordinate, we represent motion sequence $X$ with $T$ frames as $X=\{x^t\}_{t=1}^T$, where $x^t=(p_r^t, \theta_r^t, r^t)$ is SMPL~\cite{smpl} parameters, with root location $p_r^t\in \mathcal{R}^3$, root orientation $\theta_r^t \in \mathcal{R}^6$, and the orientation of $j$ limbs as $\theta^t \in \mathcal{R}^{j\times 6}$.
Based on this, the location of $j$ joints $p^t \in \mathcal{R}^{j\times 3}$ could also be computed via SMPL.
For simplicity, we use $\cdot^{-w}$ to represent the value from frame $t-w$ to $t$, and $\cdot^{+w}$ for $t$ to $t+w$.
The 6D form from \cite{rot6d} is adopted for all rotation and orientation representations.

{\bf Current Pose State} could be formulated as 
\begin{equation}
    \label{eq:xt}
    x^t=Cano(\{p_r^t, p^t, \theta_r^t, \theta^t, \dot{p}_r^t, \dot{p}^t, \dot{\theta}_r^t\}, t),
\end{equation}
where $\dot{\cdot} ^t$ indicates the first derivative at $t$. $Cano(\cdot, t)$ canonicalizes the input w.r.t. the root location and facing direction at $t$, ensuring consistency across frames.

{\bf Historical Motion State} $h^t$ captures motion from frame $t-w$ to $t$ as 
\begin{equation}
    h^t = Cano(\{p_r^{-w}, p^{-w}, \theta_r^{-w}, \dot{p}_r^{-w}, \dot{p}^{-w}, \dot{\theta}_r^{-w}\}, t),
\end{equation}
with trajectories $p_r^{-w},p^{-w}$, root orientation $\theta_r^{-w}$, velocities $\dot{p}_r^{-w},\dot{p}^{-w}$, and root angular velocity $\dot{\theta}_r^{-w}$.
For conciseness, if not specified, only the $p^{-w}$ of the hands and feet is adopted in practice.

{\bf Future Prediction} $\hat{y}^t$ is composed of two parts as $\hat{y}^t = \{\hat{x}^{t+1}, \hat{u}^t\}$.
Note that $\hat{x}^{t+1}$ is in the egocentric canonical of frame $t$, and $\hat{\dot{p}}_r,\hat{\dot{\theta}} \subset \hat{x}^{t+1}$ are adopted to transit from $t$ to $t+1$.
Mirroring the historical motion state $h^t$, we also incorporate a comprehensive set of future trajectories, facing directions and their changing rates as $\hat{u}^t = \{p_r^{+f}, p^{+f}, \theta^{+f}, \dot{p}_r^{+f}, \dot{p}^{+f}, \dot{\theta}^{+f}\}$ for a window size of $f$, all in the canonical view of $t+1$. 
Empirically, integrating $\hat{u}^t$ significantly enhances the model’s precision in navigating toward the target.

\subsection{Control Signals}
\label{sec:signals}
In this section, we introduce how the different control signals are encoded.
Note that our models can operate without these \textbf{optional} signals.

{\bf Canonical Occupancy.} 
\label{sec:occu}
A simple voxel-based representation is adopted to encode how the canonical space is occupied by the scenes.
We utilize a cubic point grid $o^t\in \{0, 1\}^{s\times s \times s}$ characterized by the grid size $s$ and the grid unit size $u$ at frame $t$. 
The grid is positioned in the human’s canonical space, with its center offset by $s/4$ in the human-facing direction. 
Each cell contains a binary value, indicating whether the cell is intersecting with the global scene occupancy $O$.
The canonical occupancy is thus represented as $c_o^t = Flatten(o^t) \in \{0,1\}^{s^3}$.
Note that because the occupancy is calculated in real-time, this control signal can naturally adapt to dynamic scenes where each timestep is treated statically. We will show in Section~\ref{sec:exp} that even though our training data is all static, our model exhibits a smooth generalization to dynamic scenes.

{\bf Target Pose} is also considered a flexible control signal, which can be given in two ways. 
The first is the positions of the root, hands, and feet, each defined in relation to the canonical space of the current pose, represented as $c^t_t \in \mathcal{R}^{5 \times 3}$. In practice, these could be assigned manually or generated from off-the-shelf models~\cite{zhao2022compositional}.
Also, partial or blank targets are acceptable.
The second is the future trajectory of end-effectors, denoted as $c_t^{+f}$ for a future window size of $f$.
This paves the way for different applications in Section~\ref{sec:exp}.

\subsection{Controller Architecture}
\label{sec:architecture}
\begin{wrapfigure}{r}{0.3\textwidth}
    \centering
    \includegraphics[width=\linewidth]{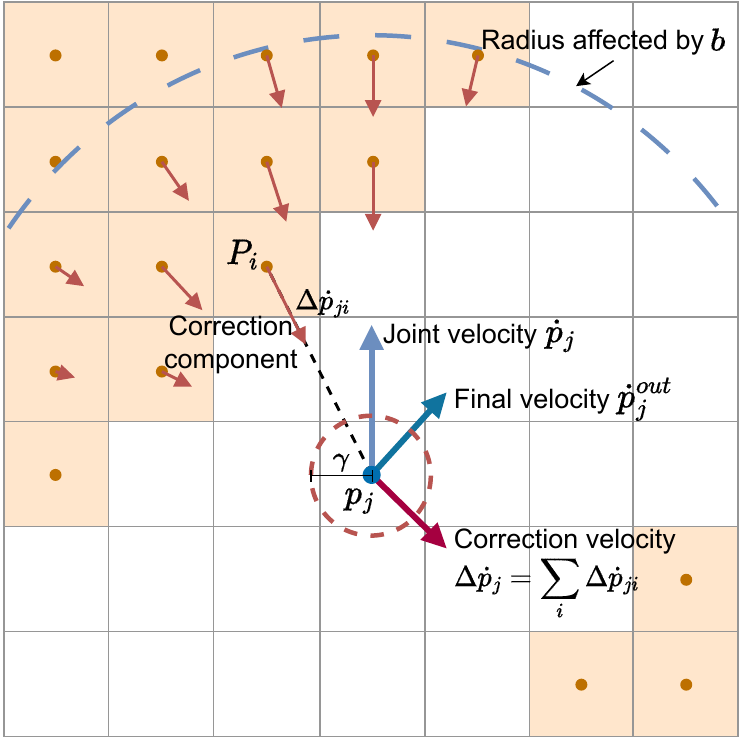}
    \caption{Occupied voxel centers $\{P_i\}$ are located at the top left of the figure. The initial velocity vector, $\dot{p}_j$, is directed upwards, posing a risk of collision. To mitigate this, the occupancy field introduces a corrective velocity component $\Delta \dot{p}_j$. This redirects the final velocity $\dot{p}_j^{out}$ to avoid collision.}
    \label{fig:field-regu}
\end{wrapfigure}
We adopt a lightweight transformer-based architecture.
$n$ fully connected layers first project the $n$ input features from $\{h^t, x^t, c^t\}$ to the latent $z \in \mathcal{R}^{n \times d}$.
$z$ is then fed into encoder $E$ to distill dynamic information from the motion states and control signals.
We devise $E$ as a sequence of Transformer encoders, which takes $z$ as input and outputs refined dynamic information $\hat{z} = E(z) \in \mathcal{R}^{n \times d}$. 
Finally, we adopt a transformer decoder $D$ followed by a linear layer to condense the dynamic information, resulting in the final predictions.
With learnable query $q \in \mathcal{R}^{d}$ and the dynamic information $\hat{z}$ serving as K and V, respectively, the output of the transformer is further concatenated with a random noise $\epsilon \in \mathcal{R}^{d}$ for randomness.
The linear layer then outputs the final prediction as $\hat{y}^t=D(\hat{z}, q, \epsilon)$.

Furthermore, to enhance collision avoidance, we introduce a novel module called \textbf{Field Regulation}.
The basic idea is that humans tend to move faster in obstacle-sparse areas and decelerate in complex environments or when an obstacle is nearby. 
At very short distances, especially with an imminent collision happening, a complete halt is necessary.
Given this and taking inspiration from the potential field in robot motion planning~\cite{hwang1992potential}, we propose \textit{Occupancy Field}, which exerts negligible influence on humans at a distance but escalates to a sturdy repulsion as proximity increases, thereby compelling the human to decelerate and avoid collision. The occupancy field influences human motion only when a human is in motion toward it. The field functions as
\begin{equation}
    \Delta \dot{p}^{t+1}_j = \mathcal{F}(\dot{p}^{t+1}_j, c_o^t). 
\end{equation}
The field $\mathcal{F}$ calculates a correction $\Delta \dot{p}^{t+1}$ w.r.t. predicted joint velocity $\dot{p}^{t+1}_j \in \{\dot{p}^{t+1}_r,\dot{p}^{t+1}\}$ and canonical occupancy $c_o^t$, which is then added to $\dot{p}^{t+1}$.
For simplicity, the superscripts indicating the time steps are omitted in the following equations.
In detail, given the center points of all the occupied voxels in $c_o$ as $P \subset \mathcal{R}^{n_p \times 3}$, the correction is calculated as 
\begin{align}
    \Delta \dot{p}_j &= \sum_{P_i \in P} - k ~ r_i~ max\left(0, \frac{1}{\|v_{ij}\|_a - \gamma} - b \right), \\
    r_i &= \dot{p}_j~max(0, \langle \dot{p}_j, v_{ij} \rangle ),
\end{align}
where $v_{ij}=P_i - p_j$ is the vector from the joint $j$ to the voxel center, $\langle \cdot, \cdot \rangle$ indicates the cosine between the two vectors, $\|\|_a$ denotes $a$-norm, and $k$ is the stiffness factor.
$\gamma$ and $b$ are thresholds to constrain the distance range of effective voxels, where $\gamma$ reflects the human body's width, and voxels within this range are discounted for stability. 
An illustration of Field Regulation is shown in Fig.~\ref{fig:field-regu}. 
The field regulation is incorporated in the training of the controller to adapt to it, avoiding potential foot sliding caused due to direct modification on velocities. 

\subsection{Training and Inference}
\label{sec:train-infer}

{\bf Training Objectives}
The controller is trained in a fully-supervised manner.
We employ a mixture of L1 and L2 losses as basic objectives. 
Besides, penetration loss and occupancy field loss are designed to hinder undesired collisions.
Specifically, L1 loss is utilized for rotation-related output values $\hat{y}^t_{\theta} \in \hat{y}^t$, while L2 loss is applied to the rest outputs $\hat{y}^t_p$. 
The mixture loss is formulated as 
\begin{equation}
    \mathcal{L}_{mix}=\|y^t_{\theta} - \hat{y}^t_{\theta}\|_1 + \|y^t_p - \hat{y}^t_p\|_2.
\end{equation}
Given the predicted joint location $\hat{p}^{t+1}$, penetration loss is computed as 
\begin{equation}
    \mathcal{L}_{pen}=\sum_{i=0}^j O(\hat{p}^{t+1}_j)\|ref(\hat{p}^{t+1}_j, O)\|_2,
\end{equation}
where $O(\cdot)$ returns 1 if the point is occupied in $O$ else 0, and $ref(\hat{p}^{t+1}_j, O)$ calculates the closest unoccupied voxel center in $O$ for joint $\hat{p}^{t+1}_j$.
Occupancy field loss has two components.
First, to deter undesired interactions, we penalize the proportion of the occupancy field effect relative to the output root velocity.
Second, observing occasional unnatural spikes in velocity, where the human figure seemingly attempts to navigate through an obstacle, we punish excessive velocities.
The occupancy field loss is formulated as
\begin{equation}
\mathcal{L}_{field} = \left(\frac{|\Delta \dot{p}|}{|~\,\dot{p}~\,|}\right)^2 + |~\dot{p}~|^2.
\end{equation}
The final loss is computed with loss coefficients $\alpha,\beta$ as
\begin{equation}
    \mathcal{L} = \mathcal{L}_{mix} + \alpha\mathcal{L}_{pen} + \beta\mathcal{L}_{field}.
\end{equation}

{\bf Auto-regressive Inference}.
We perform motion generation in an auto-regressive manner, enabling the model to adapt to \textbf{dynamic} environmental changes.
Since the model solely manipulates canonical information, we track the global root position $p_g$ and the facing direction $\theta_g$ of the human in the world coordinate system. 
Initially, we set past and current velocities $\dot{p}^0$ to zero.
Before each prediction at frame $t$, we compute the control signals (if there are any), \ie, the relative target position $c_t^t$ and the canonical space occupancy $c_o^t$, based on the current values of $p_g^t$ and $\theta_g^t$.
After the prediction, we update $p_g^t$ and $\theta_g^t$ with the model's output, along with the current pose state $c^{t+1}$.
Concurrently, we refresh the historical motion state $h^{t+1}$ to prepare for the next input.

\section{Experiments}
\label{sec:exp}

{\bf Implementation Details.}
The controller consists of a two-layer transformer encoder and a single-layer transformer decoder, both with a latent dimension of 512 and 8 heads.
The canonical occupancy grid size is set as $(25,25,25)$, with a unit size of 8 cm.
The controller operates at 10 FPS with the history and future window sizes both set as 1, which empirically brings more stable outcomes.
The controller is trained on the MOB, with 1,393 sequences excluded from training as the test set.
We train the controller for 75k iterations, with a cosine learning rate decay restart strategy.
The initial learning rate is set as 1e-4.
The control signals are randomly masked for augmentation during training.
Scheduled Sampling~\cite{bengio2015scheduled} is adopted.
The whole training takes approximately 6 hours on a single 12G NVIDIA Titan Xp.
More details are in the supplementary.

\subsection{Interacting with Occupancy from MOB}

\begin{figure}[!t]
    \centering
    \includegraphics[width=.95\linewidth]{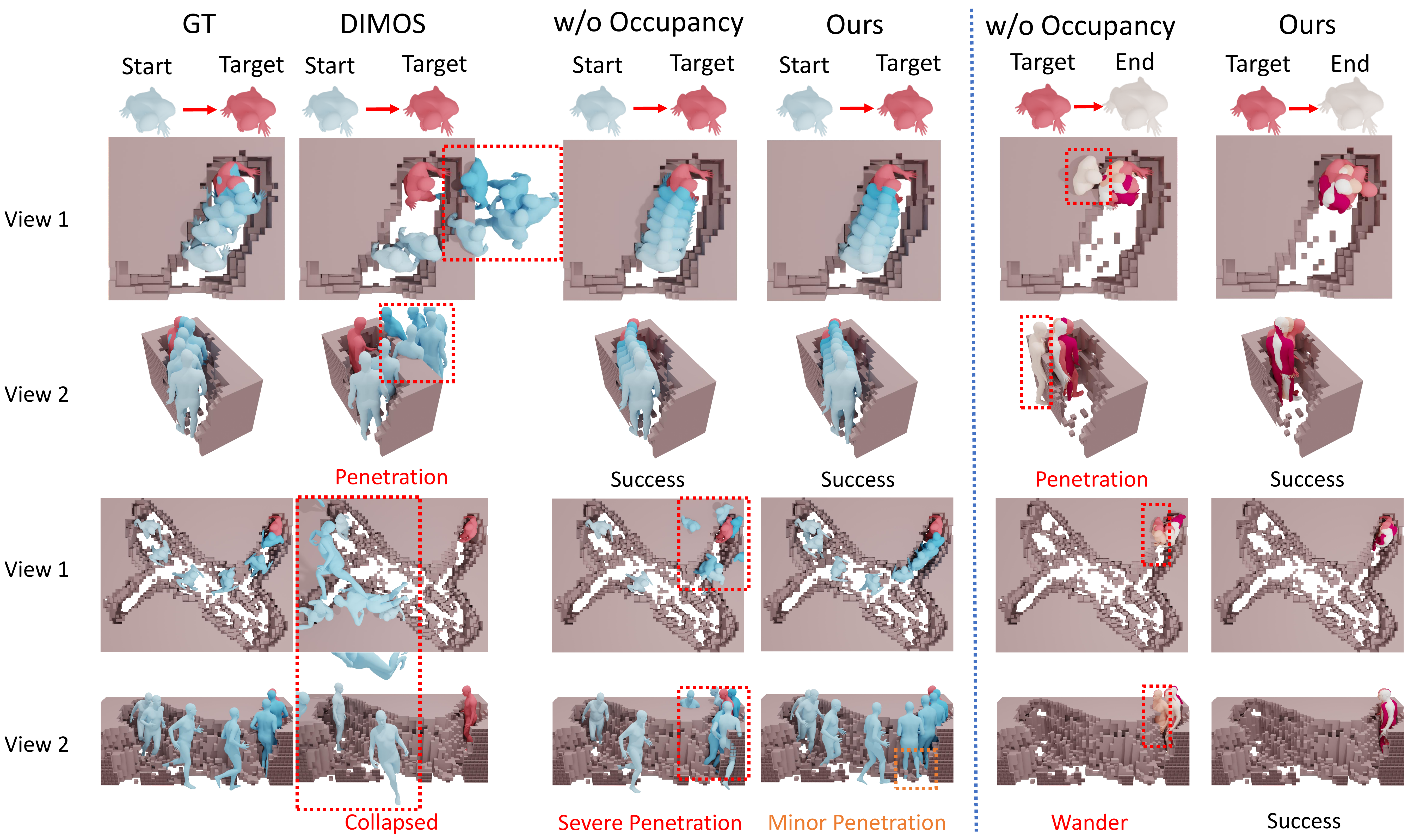}
    \caption{Our controller can naturally reach the target (white to blue) given the complex occupancy. Furthermore, it can stabilize the motions (red to white) around the targets after reaching them. Previous SOTA~\cite{zhao2023synthesizing} suffers from severe penetration and fails to reach the target. For simplicity, the ceilings of the occupancy are hidden.} 
    \label{fig:vis-human-occu-inter}
\end{figure}

\begin{table}[!t]
    \centering
    \centering
    \resizebox{.7\linewidth}{!}{\setlength{\tabcolsep}{4pt}
    \begin{tabular}{l c c c c c c}
        \hline
        Method                             & Suc. (\%) $\uparrow$ & DT (cm) $\downarrow$ & Time (s)      & FS (\%) & PEN $\downarrow$    & ERP     \\ \hline
        GT                                 & 100.00            & 0.00           & 3.00          & 3.36          & 0.00           &  -   \\ \hline
        NSM~\cite{starke2019neural}        & 11.76             & 32.79          & 5.33          & 11.04         & 134.49         & 7.26 \\
        SAMP~\cite{samp}                   & 13.94             & 52.49          & 6.54          & 9.87          & 145.74         & 6.98 \\
        DIMOS~\cite{zhao2023synthesizing}  & 15.73             & 48.97          & 6.82          & 10.52         & 123.12         & 6.82 \\ 
        Ours w/o Occupancy                      & 71.22             & \textbf{14.66} & \textbf{2.25} & \textbf{6.89} & 46.05          & 4.13 \\ 
        Ours                               & \textbf{77.91}    & 15.30          & 3.62          & 8.13          & \textbf{15.21} & 4.24 \\ \hline
        Ours w/o $L_{pen}$                 & 74.14             & 14.89          & 3.54          & 7.89          & 31.93          & 4.96 \\ 
        Ours w/o $L_{field}$               & 75.34             & 15.69          & 2.86          & 11.65         & 25.83          & 5.70 \\ 
        Ours w/o FR                        & 77.38             & 14.85          & 2.34          & \textbf{6.57} & 33.49          & 4.18 \\ 
        Ours w/ BPS                        & 73.64             & 16.83          & 3.35          & 9.34          & 36.72          & 4.72 \\ 
        \hline
    \end{tabular}}
    \captionof{table}{Quantitative results on MOB.}
    \label{tab:res-mob}
\end{table}

We first evaluate the ability to interact with occupancy volumes with a goal-reaching task on MOB.
Given an occupancy volume, a starting pose, and a target end-effector pose, the model is required to generate transition motion from the starting pose to the target pose.
We split the test set into 30k 3-second clips, adopting the first frame as the starting pose and the final frame as the target pose.
The models are required to generate motion for 10s towards the target.
There are two reasons for this setting.
First, we would like to provide enough time for the models to accomplish the tasks.
Second, we are also curious about how the models would behave after the target is reached and no new target is given, or when a very close target is given.
The results are evaluated from three aspects. 
{\bf Task Completion} is determined by whether the target pose is reached. 
The target is identified as completed if the average Euclidean distance between the generated and target pose is lower than 20 centimeters and less than 50 voxels are penetrated.
We report the success rate (Suc.).
The average minimum distance to the target pose (DT) and the time taken to reach the target (Time) are reported for reference.
Note that Time is only calculated for successful samples. 
{\bf Motion Quality} is evaluated via foot sliding (FS) and penetration (PEN) detection.
A frame is identified as FS with its foot velocity  $>$7.5 cm/s following GAMMA~\cite{zhang2022wanderings}. 
We report the percentage of frames with FS.
For PEN, we report the number of voxels penetrated per frame. 
{\bf Similarity to Dataset} is measured by the similarity between the predicted and GT target end-effector trajectory. 
We report Edit distance for the real penalty (ERP)~\cite{chen2004marriage} since the trajectories might have different lengths.
It is only given as a reference, which is not the lower the better.

{\bf Quantitative results} are shown in Tab.~\ref{tab:res-mob}. 
For DIMOS~\cite{zhao2023synthesizing}, we convert the target pose into corresponding marker sets and report the best results from different policies for each test sample.
For w/o Occ., we remove the canonical occupancy-related components.
Previous efforts frequently fail to reach the goal successfully in complex scenarios, which we will elucidate later with qualitative results. 
Even when the goal is reached, severe penetration emerges, resulting in low success rates in Tab.~\ref{tab:res-mob}.
Our models outperform DIMOS by a considerable margin with a single controller compared to multiple policies.
This reveals that the straightforward voxel-based representation is capable of encoding holistic information for human-occupancy interaction.
The model without occupancy manages to approach the target as closely and quickly as possible. However, it suffers from more severe interpenetration with the occupancy, resulting in a lower success rate.
Instead, our model with consideration of occupancy behaves more conservatively, spending more time to accomplish the tasks with less penetration.
A slightly higher foot sliding rate is also observed, which is a trade-off for the penetration-avoiding ability that field regulation brings.

\begin{figure}[!t]
    \begin{minipage}{0.5\linewidth}
        \centering
        \includegraphics[width=\linewidth]{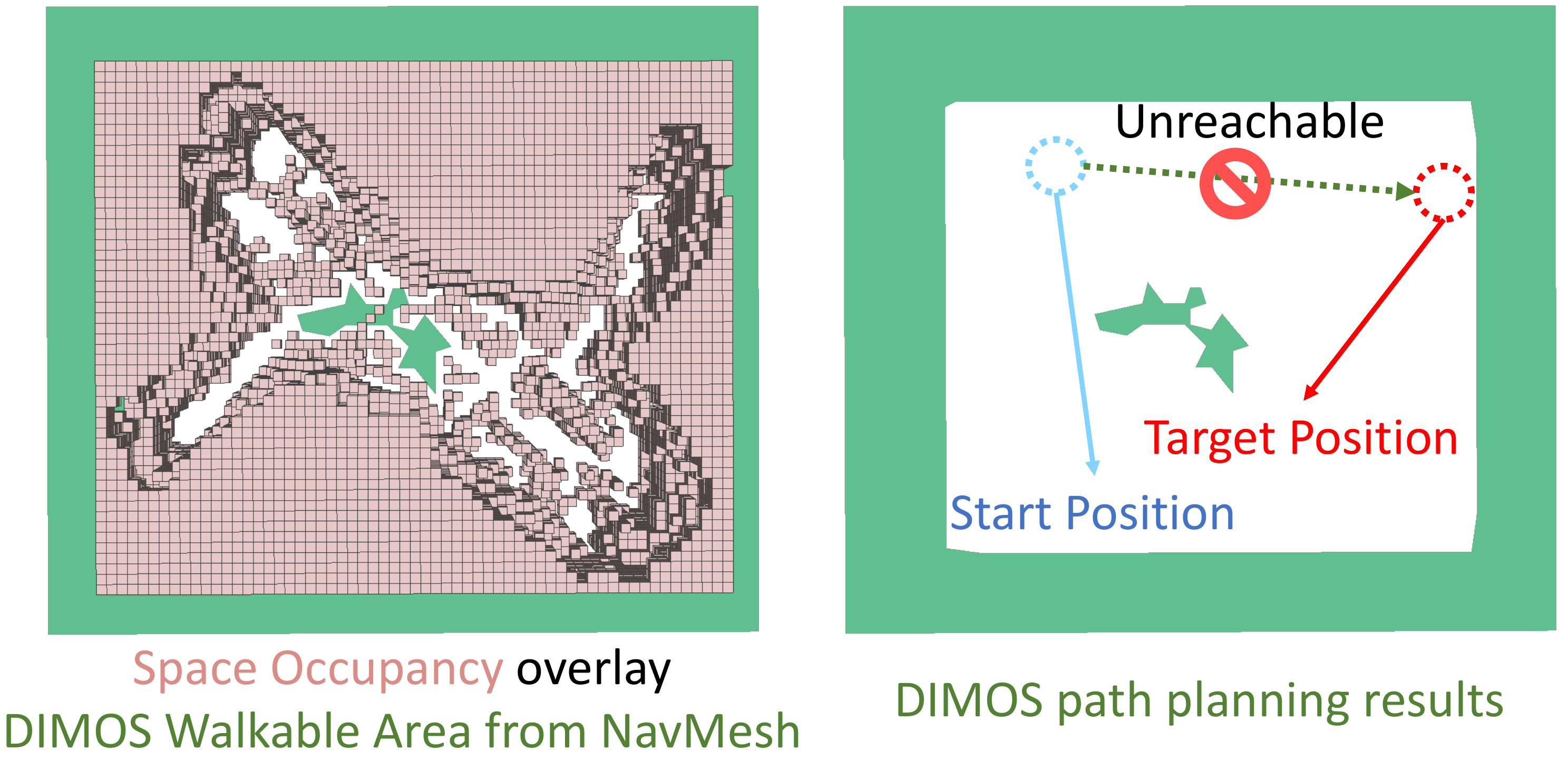}
        \captionof{figure}{A typical DIMOS~\cite{zhao2023synthesizing} failure case. The NavMesh algorithm produces sub-optimal walkable areas, resulting in path-finding failures for 69.43\% MOB samples. }
    \label{fig:nav}
    \end{minipage}
    \begin{minipage}{0.5\linewidth}
        \includegraphics[width=\linewidth]{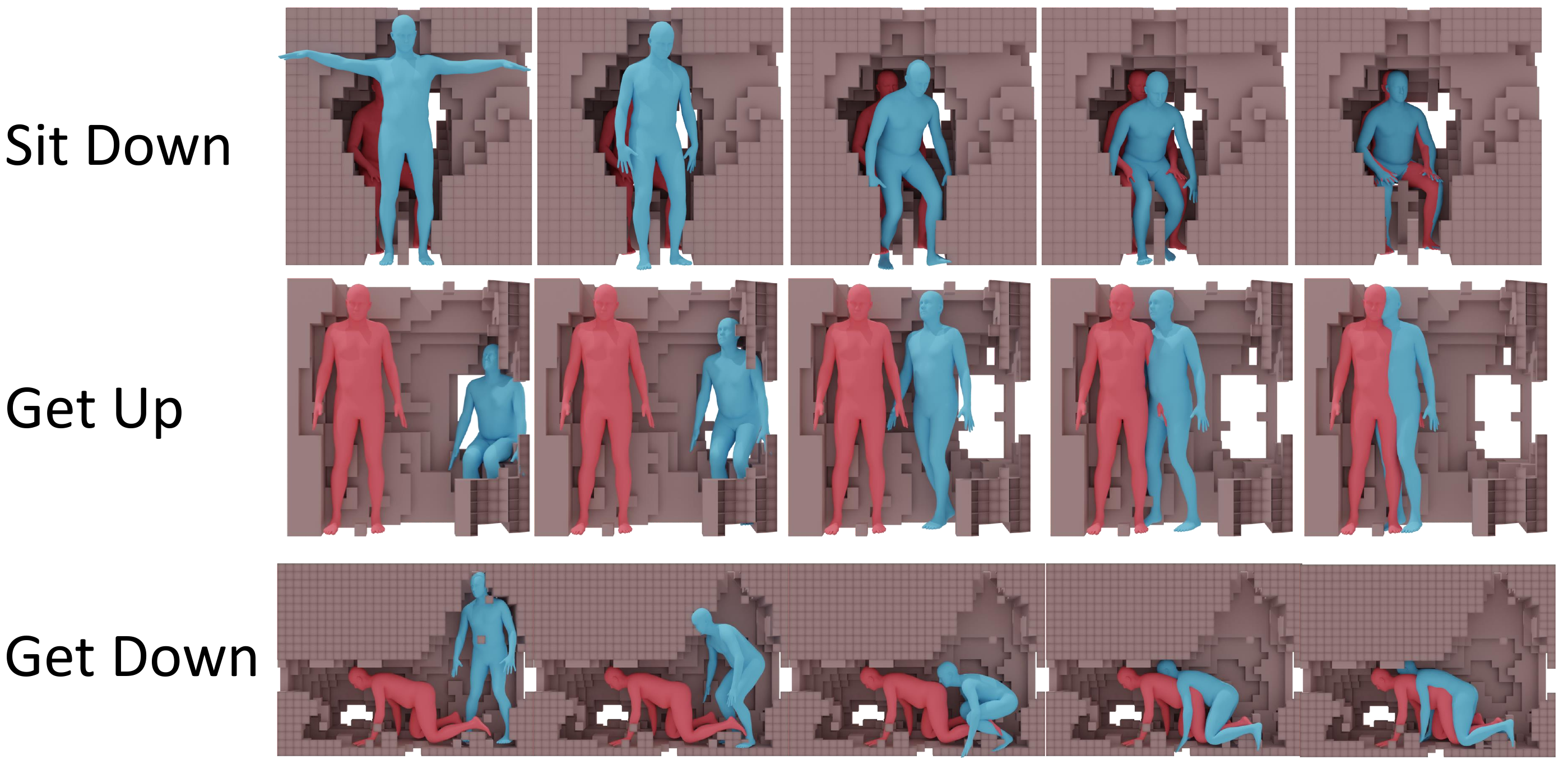}
        \caption{With the unified view of human-occupancy interaction, the same controller in Fig.~\ref{fig:vis-human-occu-inter} produces diverse motions, including sitting down, crawling, and getting up.} 
        \label{fig:sit}
    \end{minipage}
\end{figure}

{\bf Qualitative results} are shown in Fig.~\ref{fig:vis-human-occu-inter},~\ref{fig:sit}. 
Given the complex occupancy, previous state-of-the-art DIMOS~\cite{zhao2023synthesizing} frequently fails to reach the goal.
Even when it reaches the goal, severe penetration is usually observed.
Instead, our controller generates feasible motions to reach the targets.
Previous methods' typical locomotion-interaction decomposition~\cite{zhao2023synthesizing,zhang2022wanderings,wang2022towards} tends to over-simplify both.
For locomotion, most methods assume a rigid character (usually simplified as a rigid cylinder in A*~\cite{samp,wang2022towards} and NavMesh algorithms\cite{zhao2023synthesizing}),  overseeing the flexibility of humans.
It might be okay for spacious rooms, however, when posed in a cramped scene or scene with complex geometric structures like in Fig.~\ref{fig:vis-human-occu-inter}, models could fail in path-finding and thus degenerate.
As shown in Fig.~\ref{fig:nav}, for 69.43\% MOB samples, the NavMesh algorithm adopted by DIMOS failed to generate a feasible path from the start position to the target position.
The statistic for A* from SAMP~\cite{samp} is 70.29\%.
This explains the unsatisfying quantitative results of DIMOS.
For interaction, the categories are typically limited to pre-defined ones.
In contrast, with our unified view of Human-Occupancy interaction, our controller applies to various complex scenarios beyond simplified locomotion and interaction.
Furthermore, diverse motions could be performed by the same controller (Fig.~\ref{fig:sit}).
By comparing our controllers with and without occupancy, we find that training with occupancy helps to avoid penetration significantly.
Also, the controller with occupancy could stabilize the motion around the target after reaching it, while its counterpart tends to wander further inertially. 

\subsection{Interacting with Static Scenes}
Once our controller is trained on the more complex MOB, it could be effortlessly applied to realistic synthesized rooms and scene scans.

\begin{table}[!t]
    \centering
    \resizebox{.8\linewidth}{!}{\setlength{\tabcolsep}{4pt}
    \begin{tabular}{c c c c c}
        \hline
        Model & Suc. (\%) & Dist. to goal (cm) & Collision depth (cm) & FS (\%)\\
        \hline
        GT                     & 100.00 & 0.00 & 8.74  & 2.45  \\
        Ours (w/ MOB only)     & 73.01          & 7.19           & \textbf{9.56}  & 13.83 \\
        Ours (w/ CIRCLE only)  & 75.52          & 6.88           & 13.74          & \textbf{11.76} \\
        Ours (w/ MOB + CIRCLE) & \textbf{78.97} & \textbf{6.73}  & 11.85          & 12.87 \\
        \hline
    \end{tabular}}
    \caption{Results on CIRCLE~\cite{circle}.}
    \label{tab:circle}
\end{table}
\textbf{Quantitative results} is reported on CIRCLE~\cite{circle}.
Given the initial full-body pose and the target right wrist position, the model is required to generate a right-hand reaching motion.
Following \cite{circle}, success rate, distance to the goal, cumulative collision depth, and foot sliding are reported as metrics. 
Besides the controller trained on MOB (excluding CIRCLE), we also train our controller on CIRCLE-only and MOB+CIRCLE. 
We randomly split CIRCLE into a 2,565-sample train set (for the ablative baseline only) and a 453-sample test set since the original split has not been made public yet. 
The results in Tab.~\ref{tab:circle} reveal that our controller could handle \textbf{various} tasks in \textbf{complex natural} scenes well.
Also, its potential to advance HOI-related motion, like hand-reaching before grasping, is showcased.
It is noticeable that the MOB-only model demonstrates a competitive success rate and better collision avoidance, revealing the efficacy of MOB with its advanced complexity.
When incorporating MOB with CIRCLE, superior task completion is observed, showing that MOB could function as an effective auxiliary resource for the conventional HSI datasets. 

\begin{figure}[!t]
    \begin{minipage}{0.48\linewidth}
        \centering
        \includegraphics[width=.8\linewidth]{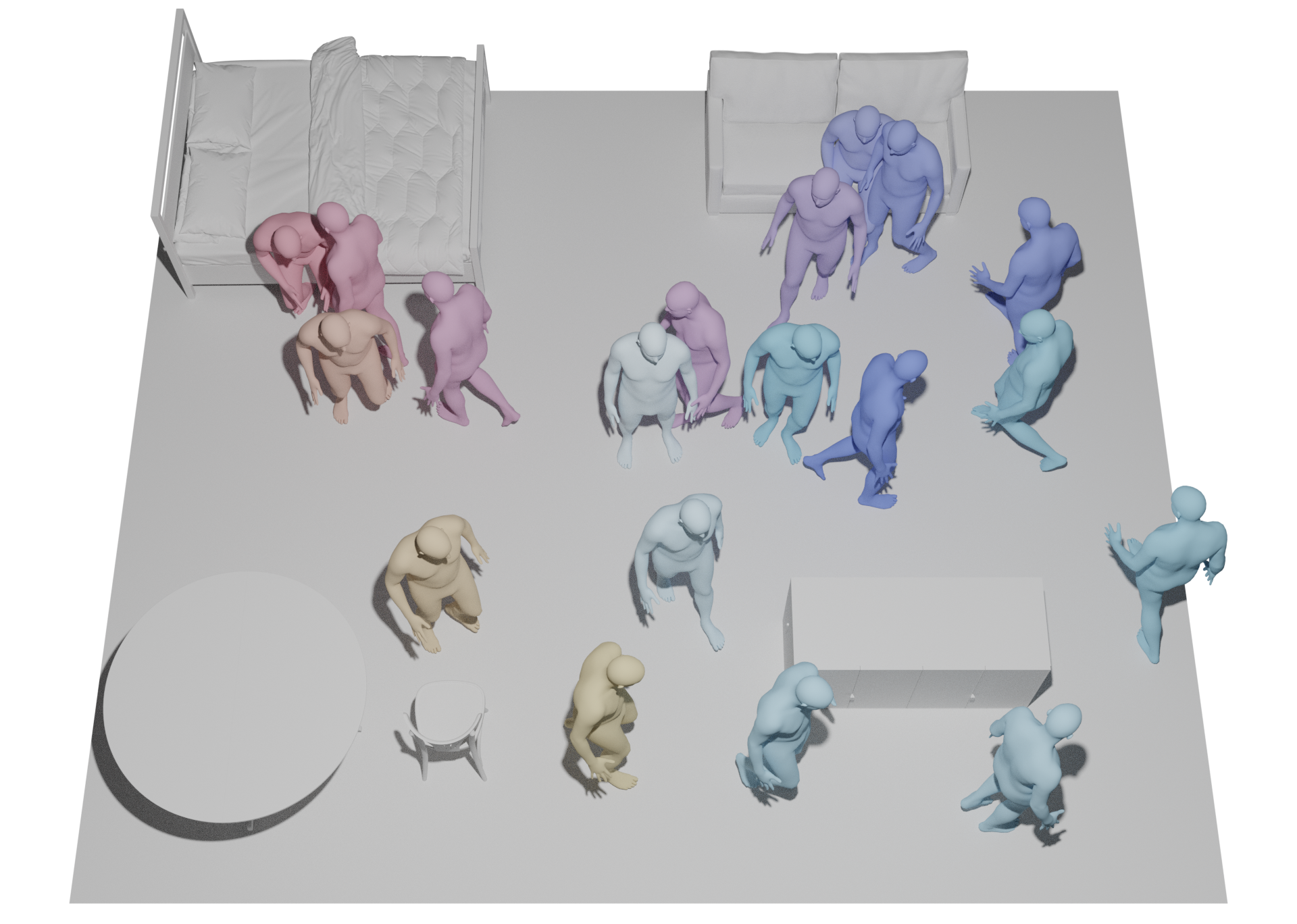}
        \caption{We can generate long-term motions in realistic rooms: A human \textcolor{cyan}{circumnavigates a cabinet}, \textcolor{blue}{sits on a sofa}, \textcolor{magenta}{perches on a bed}, and then \textcolor{orange}{resumes walking}.}
        \label{fig:long}
    \end{minipage}
    \begin{minipage}{0.52\linewidth}
        \centering
        \includegraphics[width=0.68\linewidth]{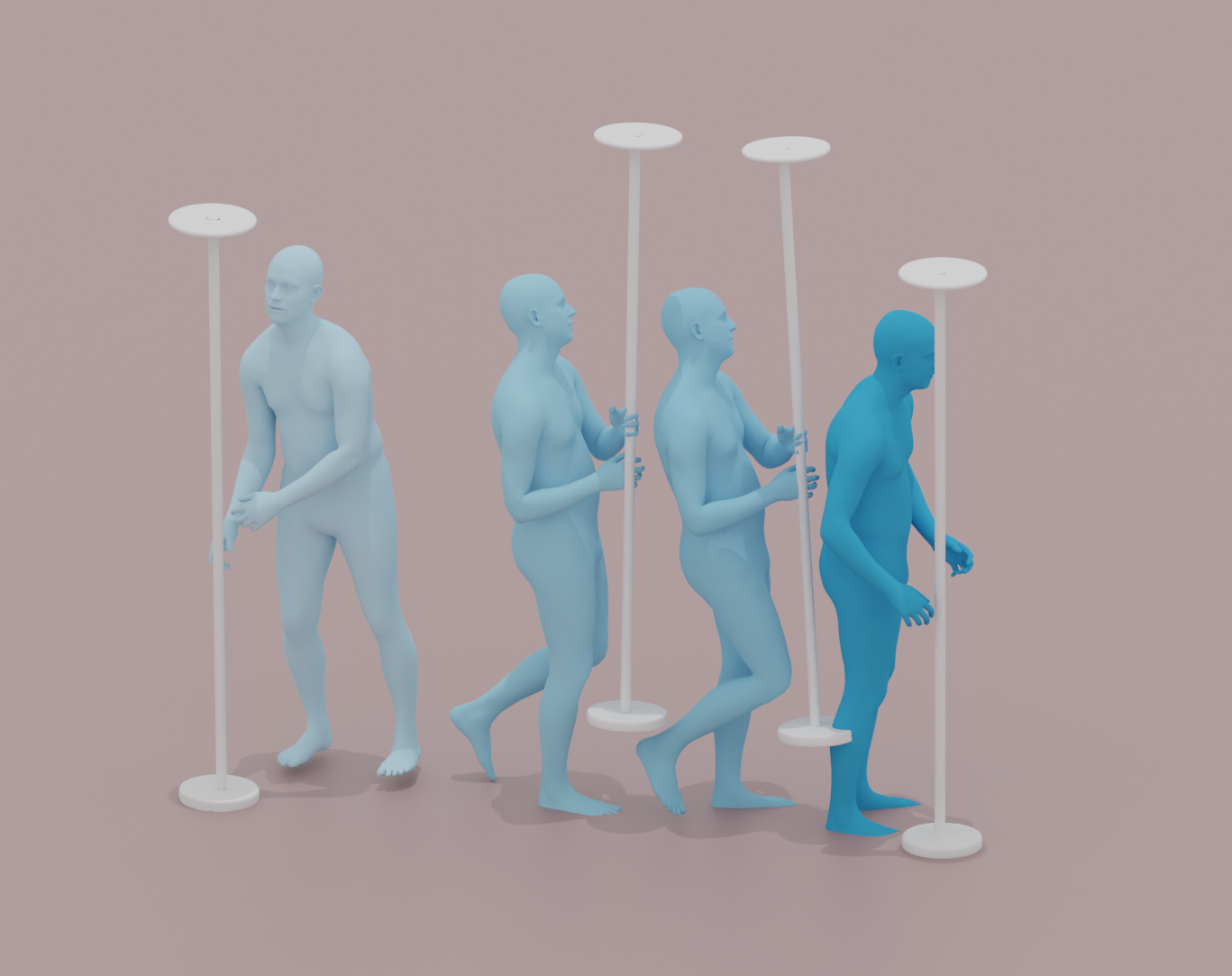}
        \caption{We synthesize Human-Object Interaction, with hand trajectory as the control signal. The human successfully moves a floor lamp to a new place.}
        \label{fig:omomo}
    \end{minipage}
\end{figure}

\textbf{Qualitative results.} We test our model in a synthetic room from DIMOS~\cite{zhao2023synthesizing} in Fig.~\ref{fig:long} and a realistic scan from Replica~\cite{straub2019replica} in Fig.~\ref{fig:1}.
The occupancy is transferred from the scene SDFs.
Our model produces convincing motions, navigates around obstacles, and transitions seamlessly between locomotion and interaction. 
Remarkably, it adapts to sitting on various objects, such as sofas, beds, and stools, revealing its versatility in recognizing different occupancies as potential seats.
The test room features sparse furnishings, highlighting the model’s ability to operate in open spaces diverging from MOB samples. 
Conversely, the Replica scene presents a more cluttered setup.
Here, the model adeptly maneuvers through tight spaces and produces plausible movements.
Furthermore, our model can generate long-term motions. 
The motions in Fig.~\ref{fig:1} and Fig.~\ref{fig:long} are generated in a single attempt by merely altering the target, both spanning around 30 seconds, which are clipped from longer motions for clear illustration in the limited space.
More results are in the supplementary materials.

\subsection{Interacting with Objects}
\label{sec:exp-hoi}
Our flexible control signal for target pose enables HOI motion synthesis. 
By fine-tuning on OMOMO~\cite{li2023object}, given the initial pose and hand trajectories, we synthesize plausible motions to interact with objects as in Figure~\ref{fig:omomo}, which exhibits rational full-body motion w.r.t. hand and object trajectories. 
More results are in the supplementary.

\subsection{Interacting with Dynamic Scenes}
\begin{figure}[!t]
    \centering
    \includegraphics[width=.9\linewidth]{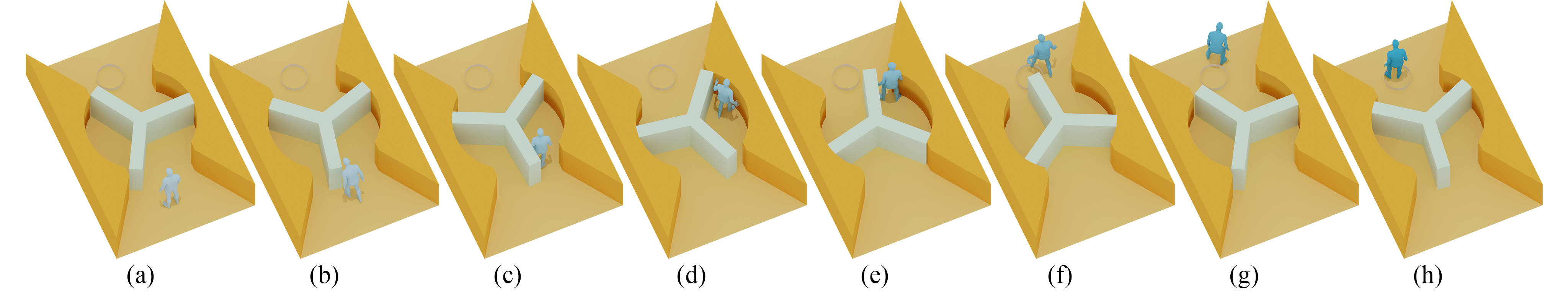}
    \caption{Our model can avoid dynamic obstacles and reasonably reach the target circle.}
    \label{fig:inter-dynamic}
\end{figure}
During training, the \textbf{canonical} occupancy perceived by our controller is dynamically changing w.r.t. to the human movement, enabling it to handle dynamic scenarios though trained on essentially static MOB.
We demonstrate the interaction in dynamic scenes with a custom automatic door scenario in Fig.~\ref{fig:inter-dynamic}. 
With the target (the circle) in front of the human~(a) blocked by a revolving door, we control the human to adapt its speed w.r.t. the door till enough clearance and halt precisely at the target.
This sequence exemplifies the model's ability to produce realistic and contextually appropriate motion in dynamic environments.
More details are in the supplementary.

\subsection{Ablation Studies}
With our controller as the baseline, selected ablation studies on MOB are reported in Tab.~\ref{tab:res-mob}.
Complete results and details are included in the supplementary.

{\bf Different loss terms.} 
By removing penetration loss $L_{pen}$, a major degeneration in Success rate and PEN is observed. 
This shows that explicitly punishing penetration is necessary. 
The removal of $L_{field}$ also hurts the success rate by 2.57\%, with PEN increasing by 10.62. 
FS also increases notably.
We identify that without $L_{field}$, the field regulation would generate unnatural spikes. 

{\bf Field Regulation (FR)} is also evaluated. 
Eliminating the field regulation decreases the success rate slightly, with a notable PEN increase (+18.28) and FS decrease (-1.56\%), revealing the tradeoff it conducts between PEN and FS. 

{\bf Occupancy representation.} 
We replaced the voxel-based representation of canonical occupancy with the more compact and efficient BPS~\cite{bps}.
The field regulation is also modified accordingly.
Degradation is observed in most metrics.
Despite its efficiency, BPS might ignore too much information and bring instability to the field regulation.

\section{Discussion}

\subsubsection{Limitations.}
\textbf{First}, when interacting with dynamic scenes, our controller could perform sub-optimally with slight collisions. This could be ascribed to the space occupancy changing rate gap between training data and testing scenes. During training, the changing rate is determined by the human locomotion speed only. While for dynamic scenes, there exists additional space occupancy speed. Mitigating this gap is worth exploring.
\textbf{Second}, foot sliding is another artifact that could happen. However, the metrics are based on simple heuristics with gaps compared to real foot sliding. A reasonable and effective tool to address this would be promising.
\textbf{Third}, the current MOB poses its main focus on static HSI. 
Despite the proposed pipeline that could apply to HOI motion synthesis, further exploration of the space-occupancy view of HOI would be an interesting future direction.
\textbf{Finally}, the proposed versatile controller only accepts target poses, space occupancy, and joint trajectories as control signals. Extending the controller for more signals would be promising.

\subsubsection{Broader Impacts.} The exploration of HSI is beneficial for navigation in complex environments, with significant impacts on Embodied AI and Robotics. Also, multiple fields like animation production and interior design could be advanced. However, this might also be misused for false information.

\section{Conclusion}
We revisited static HSI from a unified space occupancy view by identifying the major component of static HSI as Human-Occupancy Interaction.
Large-scale Human Occupancy Interaction database MOB was aggregated from motion-only data, substantially improving diversity and complexity.
Moreover, a versatile auto-regressive controller was trained on MOB for Human-Occupancy Interaction generation, which supports flexible and stable generation for static HSI under a spectrum of scenarios without training with real 3D scene data.

\section*{Acknowledgments}
This work is supported in part by the
National Natural Science Foundation of China under Grants No.62306175, No.62302296, 
National Key Research and Development Project of China (No.2022ZD0160102, No.2021ZD0110704), 
Shanghai Artificial Intelligence Laboratory, XPLORER PRIZE grants.

%
%
\bibliographystyle{splncs04}
\bibliography{main}

\begin{thebibliography}{10}
\providecommand{\url}[1]{\texttt{#1}}
\providecommand{\urlprefix}{URL }
\providecommand{\doi}[1]{https://doi.org/#1}

\bibitem{circle}
Ara{\'u}jo, J.P., Li, J., Vetrivel, K., Agarwal, R., Wu, J., Gopinath, D., Clegg, A.W., Liu, K.: Circle: Capture in rich contextual environments. In: Proceedings of the IEEE/CVF Conference on Computer Vision and Pattern Recognition. pp. 21211--21221 (2023)

\bibitem{athanasiou2022teach}
Athanasiou, N., Petrovich, M., Black, M.J., Varol, G.: Teach: Temporal action composition for 3d humans. In: 2022 International Conference on 3D Vision (3DV). pp. 414--423. IEEE (2022)

\bibitem{bengio2015scheduled}
Bengio, S., Vinyals, O., Jaitly, N., Shazeer, N.: Scheduled sampling for sequence prediction with recurrent neural networks. Advances in neural information processing systems  \textbf{28} (2015)

\bibitem{behave}
Bhatnagar, B.L., Xie, X., Petrov, I., Sminchisescu, C., Theobalt, C., Pons-Moll, G.: Behave: Dataset and method for tracking human object interactions. In: {IEEE} Conference on Computer Vision and Pattern Recognition (CVPR). {IEEE} (jun 2022)

\bibitem{humman}
Cai, Z., Ren, D., Zeng, A., Lin, Z., Yu, T., Wang, W., Fan, X., Gao, Y., Yu, Y., Pan, L., Hong, F., Zhang, M., Loy, C.C., Yang, L., Liu, Z.: Humman: Multi-modal 4d human dataset for versatile sensing and modeling. In: Avidan, S., Brostow, G., Ciss{\'e}, M., Farinella, G.M., Hassner, T. (eds.) Computer Vision -- ECCV 2022. pp. 557--577. Springer Nature Switzerland, Cham (2022)

\bibitem{gtahu}
Cai, Z., Zhang, M., Ren, J., Wei, C., Ren, D., Lin, Z., Zhao, H., Yang, L., Liu, Z.: Playing for 3d human recovery. arXiv preprint arXiv:2110.07588  (2021)

\bibitem{chen2004marriage}
Chen, L., Ng, R.: On the marriage of lp-norms and edit distance. In: Proceedings of the Thirtieth international conference on Very large data bases-Volume 30. pp. 792--803 (2004)

\bibitem{mld}
Chen, X., Jiang, B., Liu, W., Huang, Z., Fu, B., Chen, T., Yu, G.: Executing your commands via motion diffusion in latent space. In: Proceedings of the IEEE/CVF Conference on Computer Vision and Pattern Recognition. pp. 18000--18010 (2023)

\bibitem{arctic}
Fan, Z., Taheri, O., Tzionas, D., Kocabas, M., Kaufmann, M., Black, M.J., Hilliges, O.: {ARCTIC}: A dataset for dexterous bimanual hand-object manipulation. In: Proceedings IEEE Conference on Computer Vision and Pattern Recognition (CVPR) (2023)

\bibitem{chi3d}
Fieraru, M., Zanfir, M., Oneata, E., Popa, A.I., Olaru, V., Sminchisescu, C.: Three-dimensional reconstruction of human interactions. In: The IEEE/CVF Conference on Computer Vision and Pattern Recognition (CVPR) (June 2020)

\bibitem{fit3d}
Fieraru, M., Zanfir, M., Pirlea, S.C., Olaru, V., Sminchisescu, C.: Aifit: Automatic 3d human-interpretable feedback models for fitness training. In: The IEEE/CVF Conference on Computer Vision and Pattern Recognition (CVPR) (June 2021)

\bibitem{hml3d}
Guo, C., Zou, S., Zuo, X., Wang, S., Ji, W., Li, X., Cheng, L.: Generating diverse and natural 3d human motions from text. In: Proceedings of the IEEE/CVF Conference on Computer Vision and Pattern Recognition (CVPR). pp. 5152--5161 (June 2022)

\bibitem{guo2020action2motion}
Guo, C., Zuo, X., Wang, S., Zou, S., Sun, Q., Deng, A., Gong, M., Cheng, L.: Action2motion: Conditioned generation of 3d human motions. In: Proceedings of the 28th ACM International Conference on Multimedia. pp. 2021--2029 (2020)

\bibitem{guo2023back}
Guo, W., Du, Y., Shen, X., Lepetit, V., Alameda-Pineda, X., Moreno-Noguer, F.: Back to mlp: A simple baseline for human motion prediction. In: Proceedings of the IEEE/CVF Winter Conference on Applications of Computer Vision. pp. 4809--4819 (2023)

\bibitem{harvey2020robust}
Harvey, F.G., Yurick, M., Nowrouzezahrai, D., Pal, C.: Robust motion in-betweening. ACM Transactions on Graphics (TOG)  \textbf{39}(4),  60--1 (2020)

\bibitem{samp}
Hassan, M., Ceylan, D., Villegas, R., Saito, J., Yang, J., Zhou, Y., Black, M.J.: Stochastic scene-aware motion prediction. In: Proceedings of the IEEE/CVF International Conference on Computer Vision. pp. 11374--11384 (2021)

\bibitem{prox}
Hassan, M., Choutas, V., Tzionas, D., Black, M.J.: Resolving {3D} human pose ambiguities with {3D} scene constraints. In: International Conference on Computer Vision. pp. 2282--2292 (Oct 2019), \url{https://prox.is.tue.mpg.de}

\bibitem{hassan2021populating}
Hassan, M., Ghosh, P., Tesch, J., Tzionas, D., Black, M.J.: Populating 3d scenes by learning human-scene interaction. In: Proceedings of the IEEE/CVF Conference on Computer Vision and Pattern Recognition. pp. 14708--14718 (2021)

\bibitem{InterPhysHassan2023}
Hassan, M., Guo, Y., Wang, T., Black, M., Fidler, S., Peng, X.B.: Synthesizing physical character-scene interactions. In: ACM SIGGRAPH 2023 Conference Proceedings. SIGGRAPH '23, Association for Computing Machinery, New York, NY, USA (2023). \doi{10.1145/3588432.3591525}, \url{https://doi.org/10.1145/3588432.3591525}

\bibitem{hernandez2019human}
Hernandez, A., Gall, J., Moreno-Noguer, F.: Human motion prediction via spatio-temporal inpainting. In: Proceedings of the IEEE/CVF International Conference on Computer Vision. pp. 7134--7143 (2019)

\bibitem{huang2023diffusion}
Huang, S., Wang, Z., Li, P., Jia, B., Liu, T., Zhu, Y., Liang, W., Zhu, S.C.: Diffusion-based generation, optimization, and planning in 3d scenes. In: Proceedings of the IEEE/CVF Conference on Computer Vision and Pattern Recognition. pp. 16750--16761 (2023)

\bibitem{hwang1992potential}
Hwang, Y.K., Ahuja, N., et~al.: A potential field approach to path planning. IEEE transactions on robotics and automation  \textbf{8}(1),  23--32 (1992)

\bibitem{uestc}
Ji, Y., Xu, F., Yang, Y., Shen, F., Shen, H.T., Zheng, W.S.: A large-scale rgb-d database for arbitrary-view human action recognition. In: Proceedings of the 26th ACM international Conference on Multimedia. pp. 1510--1518 (2018)

\bibitem{lee2023locomotion}
Lee, J., Joo, H.: Locomotion-action-manipulation: Synthesizing human-scene interactions in complex 3d environments. In: Proceedings of the IEEE/CVF International Conference on Computer Vision (ICCV). pp. 9663--9674 (October 2023)

\bibitem{li2023object}
Li, J., Wu, J., Liu, C.K.: Object motion guided human motion synthesis. ACM Transactions on Graphics (TOG)  \textbf{42}(6),  1--11 (2023)

\bibitem{li2021ai}
Li, R., Yang, S., Ross, D.A., Kanazawa, A.: Ai choreographer: Music conditioned 3d dance generation with aist++. In: Proceedings of the IEEE/CVF International Conference on Computer Vision. pp. 13401--13412 (2021)

\bibitem{aist}
Li, R., Yang, S., Ross, D.A., Kanazawa, A.: Learn to dance with aist++: Music conditioned 3d dance generation. arXiv preprint arXiv:2101.08779  \textbf{2}(3) (2021)

\bibitem{li2020detailed}
Li, Y.L., Liu, X., Lu, H., Wang, S., Liu, J., Li, J., Lu, C.: Detailed 2d-3d joint representation for human-object interaction. In: Proceedings of the IEEE/CVF Conference on Computer Vision and Pattern Recognition. pp. 10166--10175 (2020)

\bibitem{lin2023motion}
Lin, J., Zeng, A., Lu, S., Cai, Y., Zhang, R., Wang, H., Zhang, L.: Motion-x: A large-scale 3d expressive whole-body human motion dataset. arXiv preprint arXiv:2307.00818  (2023)

\bibitem{smpl}
Loper, M., Mahmood, N., Romero, J., Pons-Moll, G., Black, M.J.: Smpl: A skinned multi-person linear model. ACM transactions on graphics (TOG)  \textbf{34}(6),  1--16 (2015)

\bibitem{posegpt}
Lucas, T., Baradel, F., Weinzaepfel, P., Rogez, G.: Posegpt: quantization-based 3d human motion generation and forecasting. In: Computer Vision--ECCV 2022: 17th European Conference, Tel Aviv, Israel, October 23--27, 2022, Proceedings, Part VI. pp. 417--435. Springer (2022)

\bibitem{amass}
Mahmood, N., Ghorbani, N., Troje, N.F., Pons-Moll, G., Black, M.J.: Amass: Archive of motion capture as surface shapes. In: Proceedings of the IEEE/CVF international conference on computer vision. pp. 5442--5451 (2019)

\bibitem{nie2022pose2room}
Nie, Y., Dai, A., Han, X., Nie{\ss}ner, M.: Pose2room: understanding 3d scenes from human activities. In: European Conference on Computer Vision. pp. 425--443. Springer (2022)

\bibitem{smplx}
Pavlakos, G., Choutas, V., Ghorbani, N., Bolkart, T., Osman, A.A.A., Tzionas, D., Black, M.J.: Expressive body capture: {3D} hands, face, and body from a single image. In: Proceedings IEEE Conf. on Computer Vision and Pattern Recognition (CVPR). pp. 10975--10985 (2019)

\bibitem{actor}
Petrovich, M., Black, M.J., Varol, G.: Action-conditioned 3d human motion synthesis with transformer vae. In: Proceedings of the IEEE/CVF International Conference on Computer Vision. pp. 10985--10995 (2021)

\bibitem{temos}
Petrovich, M., Black, M.J., Varol, G.: Temos: Generating diverse human motions from textual descriptions. In: Computer Vision--ECCV 2022: 17th European Conference, Tel Aviv, Israel, October 23--27, 2022, Proceedings, Part XXII. pp. 480--497. Springer (2022)

\bibitem{Petrovich_2023_ICCV}
Petrovich, M., Black, M.J., Varol, G.: Tmr: Text-to-motion retrieval using contrastive 3d human motion synthesis. In: Proceedings of the IEEE/CVF International Conference on Computer Vision (ICCV). pp. 9488--9497 (October 2023)

\bibitem{kit}
Plappert, M., Mandery, C., Asfour, T.: The kit motion-language dataset. Big data  \textbf{4}(4),  236--252 (2016)

\bibitem{bps}
Prokudin, S., Lassner, C., Romero, J.: Efficient learning on point clouds with basis point sets. In: Proceedings of the IEEE International Conference on Computer Vision. pp. 4332--4341 (2019)

\bibitem{puig2023habitat3}
Puig, X., Undersander, E., Szot, A., Cote, M.D., Partsey, R., Yang, J., Desai, R., Clegg, A.W., Hlavac, M., Min, T., Gervet, T., Vondru\v{s}, V., Berges, V.P., Turner, J., Maksymets, O., Kira, Z., Kalakrishnan, M., Malik, J., Chaplot, D.S., Jain, U., Batra, D., Rai, A., Mottaghi, R.: Habitat 3.0: A co-habitat for humans, avatars and robots (2023)

\bibitem{starke2019neural}
Starke, S., Zhang, H., Komura, T., Saito, J.: Neural state machine for character-scene interactions. ACM Trans. Graph.  \textbf{38}(6),  209--1 (2019)

\bibitem{straub2019replica}
Straub, J., Whelan, T., Ma, L., Chen, Y., Wijmans, E., Green, S., Engel, J.J., Mur-Artal, R., Ren, C., Verma, S., et~al.: The replica dataset: A digital replica of indoor spaces. arXiv preprint arXiv:1906.05797  (2019)

\bibitem{szot2021habitat}
Szot, A., Clegg, A., Undersander, E., Wijmans, E., Zhao, Y., Turner, J., Maestre, N., Mukadam, M., Chaplot, D., Maksymets, O., Gokaslan, A., Vondrus, V., Dharur, S., Meier, F., Galuba, W., Chang, A., Kira, Z., Koltun, V., Malik, J., Savva, M., Batra, D.: Habitat 2.0: Training home assistants to rearrange their habitat. In: Advances in Neural Information Processing Systems (NeurIPS) (2021)

\bibitem{taheri2022goal}
Taheri, O., Choutas, V., Black, M.J., Tzionas, D.: Goal: Generating 4d whole-body motion for hand-object grasping. In: Proceedings of the IEEE/CVF Conference on Computer Vision and Pattern Recognition. pp. 13263--13273 (2022)

\bibitem{grab}
Taheri, O., Ghorbani, N., Black, M.J., Tzionas, D.: {GRAB}: A dataset of whole-body human grasping of objects. In: ECCV (2020), \url{https://grab.is.tue.mpg.de}

\bibitem{tang2022flag3d}
Tang, Y., Liu, J., Liu, A., Yang, B., Dai, W., Rao, Y., Lu, J., Zhou, J., Li, X.: Flag3d: A 3d fitness activity dataset with language instruction. arXiv preprint arXiv:2212.04638  (2022)

\bibitem{tessler2023calm}
Tessler, C., Kasten, Y., Guo, Y., Mannor, S., Chechik, G., Peng, X.B.: Calm: Conditional adversarial latent models for directable virtual characters. In: ACM SIGGRAPH 2023 Conference Proceedings. SIGGRAPH '23, Association for Computing Machinery, New York, NY, USA (2023). \doi{10.1145/3588432.3591541}, \url{https://doi.org/10.1145/3588432.3591541}

\bibitem{tevet2022motionclip}
Tevet, G., Gordon, B., Hertz, A., Bermano, A.H., Cohen-Or, D.: Motionclip: Exposing human motion generation to clip space. In: Computer Vision--ECCV 2022: 17th European Conference, Tel Aviv, Israel, October 23--27, 2022, Proceedings, Part XXII. pp. 358--374. Springer (2022)

\bibitem{hmdm}
Tevet, G., Raab, S., Gordon, B., Shafir, Y., Cohen-or, D., Bermano, A.H.: Human motion diffusion model. In: The Eleventh International Conference on Learning Representations (2022)

\bibitem{wang2021synthesizing}
Wang, J., Xu, H., Xu, J., Liu, S., Wang, X.: Synthesizing long-term 3d human motion and interaction in 3d scenes. In: Proceedings of the IEEE/CVF Conference on Computer Vision and Pattern Recognition. pp. 9401--9411 (2021)

\bibitem{wang2022towards}
Wang, J., Rong, Y., Liu, J., Yan, S., Lin, D., Dai, B.: Towards diverse and natural scene-aware 3d human motion synthesis. In: Proceedings of the IEEE/CVF Conference on Computer Vision and Pattern Recognition. pp. 20460--20469 (2022)

\bibitem{humanise}
Wang, Z., Chen, Y., Liu, T., Zhu, Y., Liang, W., Huang, S.: Humanise: Language-conditioned human motion generation in 3d scenes. Advances in Neural Information Processing Systems  \textbf{35},  14959--14971 (2022)

\bibitem{wu2022saga}
Wu, Y., Wang, J., Zhang, Y., Zhang, S., Hilliges, O., Yu, F., Tang, S.: Saga: Stochastic whole-body grasping with contact. In: European Conference on Computer Vision. pp. 257--274. Springer (2022)

\bibitem{ActFormer}
Xu, L., Song, Z., Wang, D., Su, J., Fang, Z., Ding, C., Gan, W., Yan, Y., Jin, X., Yang, X., Zeng, W., Wu, W.: Actformer: A gan-based transformer towards general action-conditioned 3d human motion generation. In: Proceedings of the IEEE/CVF International Conference on Computer Vision (ICCV). pp. 2228--2238 (October 2023)

\bibitem{ye2022scene}
Ye, S., Wang, Y., Li, J., Park, D., Liu, C.K., Xu, H., Wu, J.: Scene synthesis from human motion. In: SIGGRAPH Asia 2022 Conference Papers (2022)

\bibitem{Yi_2023_CVPR}
Yi, H., Huang, C.H.P., Tripathi, S., Hering, L., Thies, J., Black, M.J.: Mime: Human-aware 3d scene generation. In: Proceedings of the IEEE/CVF Conference on Computer Vision and Pattern Recognition (CVPR). pp. 12965--12976 (June 2023)

\bibitem{yi2022human}
Yi, H., Huang, C.H.P., Tzionas, D., Kocabas, M., Hassan, M., Tang, S., Thies, J., Black, M.J.: Human-aware object placement for visual environment reconstruction. In: Proceedings of the IEEE/CVF Conference on Computer Vision and Pattern Recognition. pp. 3959--3970 (2022)

\bibitem{zhang2020perceiving}
Zhang, J.Y., Pepose, S., Joo, H., Ramanan, D., Malik, J., Kanazawa, A.: Perceiving 3d human-object spatial arrangements from a single image in the wild. In: Computer Vision--ECCV 2020: 16th European Conference, Glasgow, UK, August 23--28, 2020, Proceedings, Part XII 16. pp. 34--51. Springer (2020)

\bibitem{t2mgpt}
Zhang, J., Zhang, Y., Cun, X., Zhang, Y., Zhao, H., Lu, H., Shen, X., Shan, Y.: Generating human motion from textual descriptions with discrete representations. In: Proceedings of the IEEE/CVF Conference on Computer Vision and Pattern Recognition (CVPR). pp. 14730--14740 (June 2023)

\bibitem{motiondiffuse}
Zhang, M., Cai, Z., Pan, L., Hong, F., Guo, X., Yang, L., Liu, Z.: Motiondiffuse: Text-driven human motion generation with diffusion model. arXiv preprint arXiv:2208.15001  (2022)

\bibitem{ReMoDiffuse}
Zhang, M., Guo, X., Pan, L., Cai, Z., Hong, F., Li, H., Yang, L., Liu, Z.: Remodiffuse: Retrieval-augmented motion diffusion model. In: Proceedings of the IEEE/CVF International Conference on Computer Vision (ICCV). pp. 364--373 (October 2023)

\bibitem{zhang2022egobody}
Zhang, S., Ma, Q., Zhang, Y., Qian, Z., Kwon, T., Pollefeys, M., Bogo, F., Tang, S.: Egobody: Human body shape and motion of interacting people from head-mounted devices. In: European Conference on Computer Vision. pp. 180--200. Springer (2022)

\bibitem{zhang2020place}
Zhang, S., Zhang, Y., Ma, Q., Black, M.J., Tang, S.: Place: Proximity learning of articulation and contact in 3d environments. In: 2020 International Conference on 3D Vision (3DV). pp. 642--651. IEEE (2020)

\bibitem{couch}
Zhang, X., Bhatnagar, B.L., Starke, S., Guzov, V., Pons-Moll, G.: Couch: Towards controllable human-chair interactions. In: European Conference on Computer Vision. pp. 518--535. Springer (2022)

\bibitem{zhang2020generating}
Zhang, Y., Hassan, M., Neumann, H., Black, M.J., Tang, S.: Generating 3d people in scenes without people. In: Proceedings of the IEEE/CVF conference on computer vision and pattern recognition. pp. 6194--6204 (2020)

\bibitem{zhang2022wanderings}
Zhang, Y., Tang, S.: The wanderings of odysseus in 3d scenes. In: Proceedings of the IEEE/CVF Conference on Computer Vision and Pattern Recognition. pp. 20481--20491 (2022)

\bibitem{zhao2022compositional}
Zhao, K., Wang, S., Zhang, Y., Beeler, T., Tang, S.: Compositional human-scene interaction synthesis with semantic control. In: European Conference on Computer Vision. pp. 311--327. Springer (2022)

\bibitem{zhao2023synthesizing}
Zhao, K., Zhang, Y., Wang, S., Beeler, T., Tang, S.: Synthesizing diverse human motions in 3d indoor scenes. In: Proceedings of the IEEE/CVF International Conference on Computer Vision (ICCV). pp. 14738--14749 (October 2023)

\bibitem{gimo}
Zheng, Y., Yang, Y., Mo, K., Li, J., Yu, T., Liu, Y., Liu, C.K., Guibas, L.J.: Gimo: Gaze-informed human motion prediction in context. In: European Conference on Computer Vision. pp. 676--694. Springer (2022)

\bibitem{rot6d}
Zhou, Y., Barnes, C., Lu, J., Yang, J., Li, H.: On the continuity of rotation representations in neural networks. In: Proceedings of the IEEE/CVF Conference on Computer Vision and Pattern Recognition. pp. 5745--5753 (2019)

\end{thebibliography}

\appendix
\section{Motion Occupancy Base Details}
\label{sec:mob-detail}

MOB aggregates 13 datasets, resulting in 98k instances as shown in Tab.~\ref{tab:mob-stat}. Note that for all the datasets, we only preserve the motion data.

\begin{table}[!ht]
    \centering
    \begin{tabular}{l c}
        \hline
        Dataset                       & \#Motion  \\
        \hline
        GTAHuman~\cite{gtahu}*        & 40k       \\
        AMASS~\cite{amass}            & 17.9k     \\
        FLAG3D~\cite{tang2022flag3d}* & 14.4k     \\
        CIRCLE~\cite{circle}*\dag     & 14k      \\
        Fit3D~\cite{fit3d}*           & 4.2k      \\
        GRAB~\cite{grab}*\dag         & 2.7k     \\
        AIST++~\cite{aist}*           & 1.4k      \\
        ARCTIC~\cite{arctic}*\dag     & 1.2k     \\ 
        CHI3D~\cite{chi3d}*           & 0.7k      \\
        BEHAVE~\cite{behave}*\dag     & 0.6k     \\
        EgoBody~\cite{zhang2022egobody}* & 0.5k   \\
        SAMP~\cite{samp}*\dag         & 0.3k     \\
        PROXD~\cite{prox}*\dag        & 0.1k     \\
        \hline
        Sum                           & 98k          \\
        \hline
    \end{tabular}
    \caption{Motion Occupancy Base statistics. $*$ means mirror augmentation is adopted for data boosting. \dag means the original dataset contains either scene or object information, but we do not use them.}
    \label{tab:mob-stat}
\end{table}

\section{Versatile Motion Controller Details}
\label{sec:con-detail}
In this section, we describe how we define the canonical space of a human.
The coordinate system adopted in our work adheres to the right-hand rule.
Following~\cite{hml3d}, we establish the human's ``right direction'' by calculating the vector sum that extends from the left shoulder to the right shoulder and from the left hip to the right hip.
Subsequently, the ``forward direction'' of the human is determined as the perpendicular vector to this ``right direction'' counterclockwise within the XY-plane.
This approach facilitates an accurate and consistent orientation mapping for the human.

\section{Experiment Details}
\label{sec:exp-detail}
\subsection{Implementation Details}
The controller consists of a two-layer transformer encoder and a single-layer transformer decoder, both with a latent dimension of 512 and 8 heads. 
The canonical occupancy grid size is set as $(25,25,25)$, with a unit size of 8 cm.
For the BPS-based representation, the size is set as 1,024.
The window sizes for the history and the future are both set as 1, which empirically brings more stable outcomes.
The binary foot contact labels are included in the current pose, indicating whether the human's feet are less than 5 cm to the ground, which slightly reduces foot skating.
The controller operates at 10 FPS with the history and future window sizes both set as 1, which empirically brings more stable outcomes.
The controller is trained on the MOB, with 1,393 sequences excluded from training as the test set. 
The training process spans 75k iterations, employing a cosine learning rate decay restart strategy, starting from an initial rate of 1e-4. 
The loss coefficients are set as $\alpha=2,\beta=1$.
For the Scheduled Sampling training strategy, we replace a subset of the inputs with the model's outputs from the preceding timestamp, with the replacement probability $p$ gradually increasing from 0 to 0.8 over the first 30 epochs. This procedure necessitates multi-step predictions at each training iteration, with our empirical choice being 4 steps.
The whole training takes approximately 6 hours on a single 12G NVIDIA Titan Xp.

\subsection{Interaction with Objects}

\begin{table}[!t]
    \centering
    \resizebox{.5\linewidth}{!}{\setlength{\tabcolsep}{3.5pt}
    \begin{tabular}{lcccc}
        \toprule
        Methods                     & Hand JPE          & MPJPE   & T$_root$       & O$_root$  \\
        \hline
        OMOMO~\cite{li2023object}   & \textbf{24.01}    & \textbf{12.42}  & 18.44         & 0.50  \\
        Ours                        & 33.27             & 16.68           & \textbf{8.83} & \textbf{0.17}    \\
        \bottomrule
    \end{tabular}}
    \caption{Quantitative results of Human-Object Interaction.}
    \label{tab:hoi}
\end{table}

Here, we set the control signal of our model to be the future trajectory of human-object contact points, with a window size of 10 timesteps. The contact points can be generated using off-the-shelf models, such as the first stage of OMOMO~\cite{li2023object}.
We fine-tune our model on the OMOMO dataset for 75k iterations and follow the first setting in OMOMO~\cite{li2023object}, where the 15 objects are used for both training and testing.
We use the same metrics of OMOMO and run only once to calculate the metrics. The quantitative results are reported in Tab.~\ref{tab:hoi}. 
Our root translation and root rotation error are significantly lower than OMOMO, indicating that our model could precisely reason the desired global path given hand trajectories. 
However, for Hand JPE and MPJPE, OMOMO provides better performance. 
This could be attributed to the noticeable model capacity gap between our lightweight real-time model and the diffusion model in OMOMO.
We provide the visualization in \texttt{6630.mp4} available on the website.

\subsection{Interaction with Dynamic Scenes}
The revolving-door scene in the main text is demonstrated, which features a circular, automatic revolving door with a radius of 1.5 m, providing space for entry and exit. This door assembly comprises three individual doors, each with a thickness of 35 cm. The doors are engineered to rotate around the Z-axis at a consistent speed of 15 degrees per second. 
At each frame, the space occupancy and target pose canonicalized w.r.t. human movement are fed into the controller.

Moreover, a customized dynamic scene involving sudden scene changes is also presented, demonstrating the ability to handle abrupt scene alterations.
The visualizations are available in \texttt{6630.mp4} available on the website.

\subsection{Ablation Studies}
For BPS-based~\cite{bps} representation, a basis point set with 1,024 points is randomly sampled in a ball centered at the pelvis with a radius of 1m. The occupancy field is also considered upon the 1,024 points.
Visualizations of ablation studies are included in \texttt{6630.mp4} available on the website.

\section{Extended Discussion}

\subsection{Field Regulation}

\begin{figure}[!h]
    \centering
    \includegraphics[width=\linewidth]{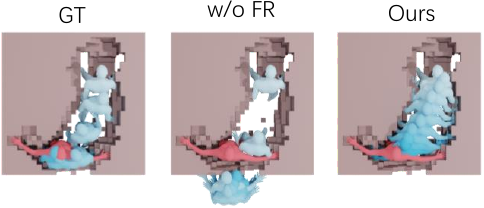}
    \caption{Additional ablative visualization for FR.}
    \label{fig:frabl}
\end{figure}

The field regulation is effective in preventing penetration, as shown in Fig.~\ref{fig:frabl}. Although it may cause slightly more severe foot sliding, the compromise is marginal (from 6.57\% to 8.13\%), as reflected in main text Tab.~{1}. Also, the weight of the computed $\Delta \dot{p}$ could be adjusted to balance foot sliding and penetration. For example, a weight of 0.1 results in 77.63\% Suc., 28.96 PEN, and 7.49 FS.

\subsection{Diversity of Generation}
\begin{figure}
    \centering
    \includegraphics[width=\linewidth]{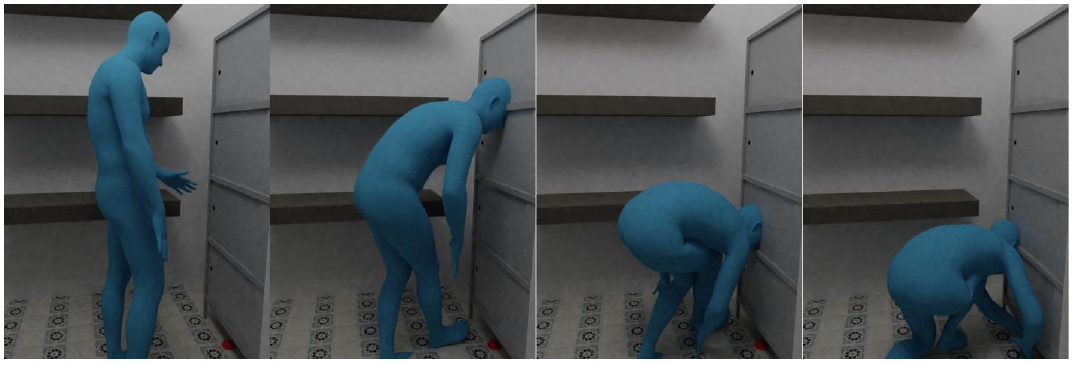}
    \caption{One of our samples on CIRCLE. The head penetration is partly due to the gap between SMPL joints and surface vertices.}
    \label{fig:div}
\end{figure}

Diversity in general scenes could be achieved with diverse target poses, as validated by main text Tab.~2 with results on CIRCLE, which is composed of diverse hand-reaching motions as shown in Fig.~\ref{fig:div}.
We can adapt to diverse target poses, resulting in diverse motion. 
Also, pseudo-occupancy is randomly added around the human in each frame, to serve as regularization and generate more diverse motions.

\begin{figure}[!t]
    \centering
    \includegraphics[width=\linewidth]{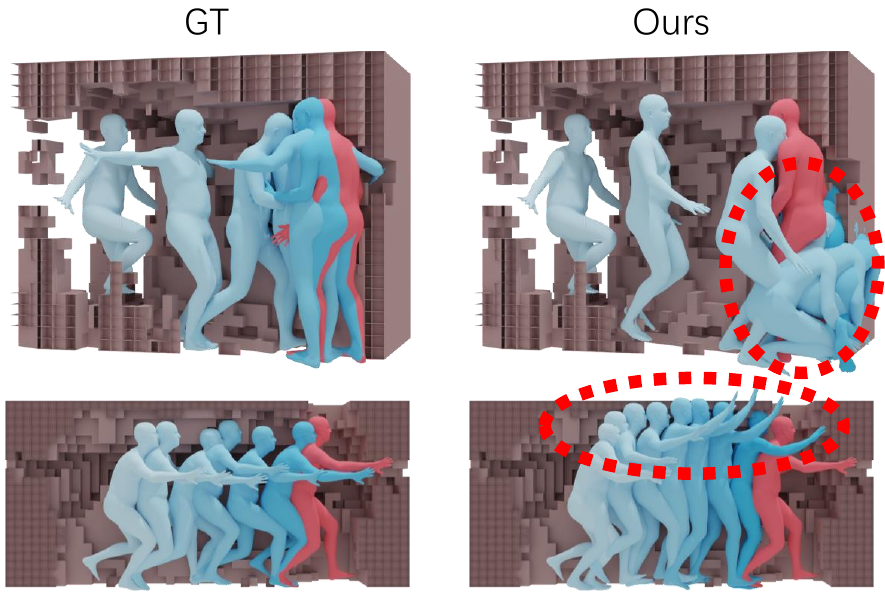}
    \caption{Failure cases.}
    \label{fig:fail}
\end{figure}

\subsection{Failure Cases}
Fig.~\ref{fig:fail} includes two cases with some occupancy carved.
Our method could be challenged when turning around in a narrow space is required as in the above case.
Controlling upper limb movements in strictly restricted space could also be hard, as our controller fails to maintain the arms at a proper height to fit in the narrow gap.

\end{document}